\pdfoutput=1

\documentclass[10pt,twocolumn,letterpaper]{article}

\usepackage{cvpr}

\usepackage[utf8]{inputenc} 
\usepackage[T1]{fontenc}    
\usepackage{url}            
\usepackage{booktabs}       
\usepackage{amsfonts}       
\usepackage{nicefrac}       
\usepackage{microtype}      
\usepackage{amsmath}
\usepackage{bbm}
\usepackage{epsfig}
\usepackage{graphicx}
\usepackage{subcaption}
\usepackage{appendix}
\usepackage[nocompress]{cite}

\cvprfinalcopy

\def\IntPhys{IntPhys 2019}

\title{\IntPhys{}: A Benchmark for Visual Intuitive Physics Understanding}

\author{Ronan~Riochet\\
Ecole~Normale~Sup\'erieur,~INRIA\\
{\tt\small ronan.riochet@inria.fr}
\and
Mario~Ynocente~Castro\\
\and
Mathieu~Bernard\\
Ecole~Normale~Sup\'erieur,~INRIA\\
\and
Adam~Lerer\\
Facebook~AI~Research\\
\and
Rob~Fergus\\
Facebook~AI~Research\\
\and
V\'eronique~Izard\\
CNRS/Universit\'e~Paris~Descartes\\
\and
Emmanuel~Dupoux\\
CoML,~ENS/CNRS/EHESS/INRIA/PSL~Research~University,~Facebook~AI~Research\\
}

\begin{document}

\maketitle

\begin{abstract}
In order to reach human performance on complex visual tasks, artificial systems need to incorporate a significant amount of understanding of the world in terms of macroscopic objects, movements, forces, etc. Inspired by work on intuitive physics in infants, we propose an evaluation benchmark which diagnoses how much a given system understands about physics by testing whether it can tell apart well matched videos of possible versus impossible events constructed with a game engine. The test requires systems to compute a physical plausibility score over an entire video. It is free of bias and can test a range of basic physical reasoning concepts. We then describe two Deep Neural Networks systems  aimed at learning intuitive physics in an unsupervised way, using only physically possible videos. The systems are trained with a future semantic mask prediction objective and tested on the possible versus impossible discrimination task. The analysis of their results compared to human data gives novel insights in the potentials and limitations of next frame prediction architectures.

\end{abstract}

\section{Introduction}

Despite impressive progress in machine vision on many tasks (face recognition \cite{wright2009robust}, object recognition \cite{NIPS2012_4824,he2016deep}, object segmentation \cite{pinheiro2015learning}, etc.), artificial systems are still far from human performance when it comes to common sense reasoning about objects in the world or understanding of complex visual scenes. Indeed, even very young children have the ability to represent macroscopic objects and track their interactions through time and space. Just a few days after birth, infants can parse their visual inputs into solid objects \cite{doi:10.1111/j.1467-8624.2006.00975.x}. At 2-4 months, they understand object permanence, and recognize that objects should follow spatio-temporally continuous trajectories \cite{kellman1983perception,spelke1995spatiotemporal}. At 6 months, they understand the notion of stability, support and causality \cite{saxe2006perception,baillargeon1992development,baillargeon1990top}. Between 8 and 10 months, they grasp the notions of gravity, inertia, and conservation of momentum in collision; between 10 and 12 months, shape constancy \cite{xu_1996}, and so on. Reverse engineering the capacity to autonomously learn and exploit intuitive physical knowledge would help building more robust and adaptable real life  applications (self-driving cars, workplace or household robots). 

\begin{figure}[t]
\begin{center}
   \includegraphics[width=0.9\linewidth]{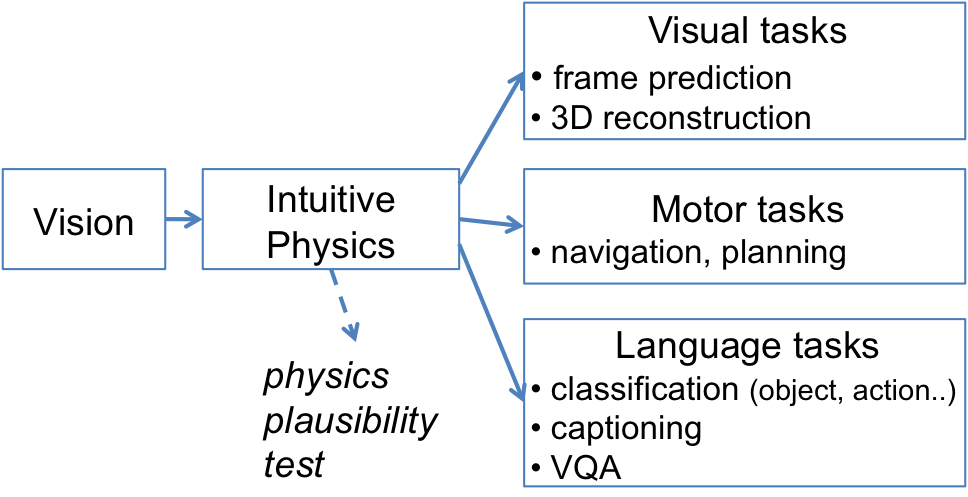}
\end{center}
   \caption{\small{Popular end-to-end applications involving scene understanding and proposed evaluation method based on physical plausibility judgments. 'Visual' tasks aim at recovering high level structure from low level (pixel) information: for instance, recovering 3D structure from static or dynamic images (e.g.,  \cite{chang2015shapenet,choy20163d}) or tracking objects  (e.g., \cite{Kristan2016b, bertinetto2016fully}). 'Motor' tasks aim at predicting the visual outcome of particular actions (e.g., \cite{finn2016unsupervised}) or to plan an action in order to reach a given outcome (e.g. \cite{Oh2015}). 'Language tasks' requires the artificial system to translate input pixels into a verbal description, either through captioning \cite{farhadi2010every} or visual question answering (VQA \cite{zitnick2013bringing}). All of these tasks involve indirectly some notion of intuitive physics. Our proposed test directly measures physical understanding in a task- and model-agnostic way.
   }}
\label{fig:tasks}
\end{figure}

Although very diverse vision tasks could benefit from some understanding of the physical world (see Figure  \ref{fig:tasks}), model of intuitive physics has been mostly developed through some form of future prediction task \cite{}. Being presented with inputs that can be pictures, video clips or actions to be performed in the case of a robot, the task is to predict future states of these input variables.  Future prediction objectives have a lot of appeal because there is no need for human annotations, and abundant data can be collected easily. The flip side is that it is difficult to find the right metric to evaluate these systems. Even though pixel-wise prediction error can be a good loss function, it is not particularly interpretable,  depends on the scale and resolution of the sensors making cross datasets comparison difficult, may not even rank the systems in a useful way: a good physics model could predict well the position of objects, but fail to reconstruct the color or texture of objects. In addition, even though the laws of macroscopic physics are deterministic, in practice many outcomes are stochastic (this is why people play dice). In other words, the outcome of any interaction between object is a distribution of object positions, making the evaluation problem even harder.  
 
Here, we propose to use an evaluation method which escapes these problems by using the prediction error not directly as a metric, but indirectly as informing a forced choice between two categories of events: \textit{possible} versus \textit{impossible events}. The intuition is the following. If a model has learned the laws of physics, it should be able to predict relatively accurately the future in video clips that show possible events, even if these videos are entirely novel. However, the model should give large prediction errors when some unlikely or impossible event happens. In other words, impossible events have a zero probability in the real world, so a more trained only with possible events should be able to generalize to other possible events, while rejecting impossible ones.

This is directly inspired by  the "violation of expectation" (VOE) paradigm in cognitive psychology, whereby infants or animals are presented with real or virtual animated 3D scenes which may contain a physical impossibility. The "surprise" reaction to the physical impossibility is measured through looking time or other physiological measures, and is taken to reflect a violation of it's internal predictions \cite{baillargeon1985object}. Similarly, our evaluation requires systems to output a scalar variable upon the presentation of a video clip, which we will call a '\textit{plausibility score}' (it could be a log probability, an inverse reconstruction error, etc). We expect the plausibility score to be lower for clips containing the violation of a physical principle than for matched clips with no violation. By varying the nature of the physical violation, one can probe different types of physical laws (conservation of objects properties, objects movement, etc.).\footnote{This has a direct parallel in 'black box' evaluation of language models in NLP. Language models are typically trained with a future prediction objective (predicting future characters or words conditioned on past ones). However, instead of evaluating theses models directly on the loss function or derivatives like perplexity, an emerging research direction is to the models on artificially constructed sentences that violate certain grammatical rules (like number agreement) measure the ability of the system to detect these violations \cite{linzen_2016}.}

As in infant's experiments, our tests are constructed in well matched sets of clips, i.e., the possible versus impossible clips differ minimally, in order to minimize the possibility of dataset biases, but are quite varied, to maximize the difficulty of solving the test through simple heuristics. Three additional advantages of this method are that (1) they provide directly interpretable results (as opposed to a prediction error, or a composite score reflecting an entire pipeline), (2) they enable to probe generalization for difficult cases outside of the training distribution, which is useful for systems that are intended to work in the real world, and (3) they enable for rigorous human-machine comparison, which is important in order to quantify how far are artificial system in matching human intuitive physical understanding. 

Our tests have also limits, which are the flip side of their advantage: They measure intuitive physics as looked through the prediction errors of a system, but do not measure how well a system might be able to use this kind of understanding. For instance, an end-to-end VQA system may have superb physical understanding (as measured by VOE) but fail miserably in connecting it with language. In this sense, VOE should be viewed as a diagnostic tool, a kind if \textit{unit testing} for physics that needs to be combined with other measures to fully evaluate end-to-end systems. Similarly these tests do not exhaustively probe for all aspects of intuitive physics, but rather break it down into a small set of basic concepts tested one at a time. Here again, unit testing does not guarantee that an entire system will work correctly, but it helps to understand what happens when it does not.

This paper is structured as follows. In Section \ref{sec:bench}, we present the \IntPhys{} Benchmark, which tests for 3 basic concepts of intuitive physics in a VOE paradigm. In Section \ref{sec:baseline}, we describe two baseline systems which are trained with a self-supervised frame prediction objective on the training set, and in Section \ref{sec:human} we analyse their performance compared to that of humans participants. In Section \ref{sec:related} we present related work and conclude in Section \ref{sec:conc} by discussing the next steps in extending this approach to more intuitive physics concepts and how they could be augmented to incorporate testing of decision and planning.

\section{Structure of the \IntPhys{} benchmark}\label{sec:bench}

\IntPhys{} is a benchmark designed to address the evaluation challenges for intuitive physics in vision systems. It can be run on any of machine vision system (captioning and VQA systems, systems performing 3D reconstruction, tracking, planning, etc), be they engineered by hand or trained using statistical learning, the only requirement being that the tested system should output a scalar for each test video clip reflecting the \textit{plausibility} of the clip as a whole. Such a score can be derived from prediction errors, or posterior probabilities, depending on the system.

In this release, which will be the first evaluation of the DARPA Machine Common Sense project, we have implemented tests for three basic concepts of the physics of macroscopic solid objects: object permanence, shape constancy, spatio-temporal continuity. Each of these concepts are tested in a series of controlled possible and impossible clips, which are presented without labels, and for which models have to return a plausibility score. The evaluation is done upon submission of these scores in CodaLab, and the results are automatically presented in a leaderboard. This benchmark also contains a training set of videos with random object interactions, in a similar environment as for the test set. This can be used either to train predictive systems or to conduct domain adaptation for systems trained on other datasets (live videos, virtual environments, robots). Obviously, the training set only contains physically possible events. 

\subsection{Three basic concepts of intuitive physics}
 \begin{figure}[t]
\begin{center}
   \includegraphics[width=1.0\linewidth]{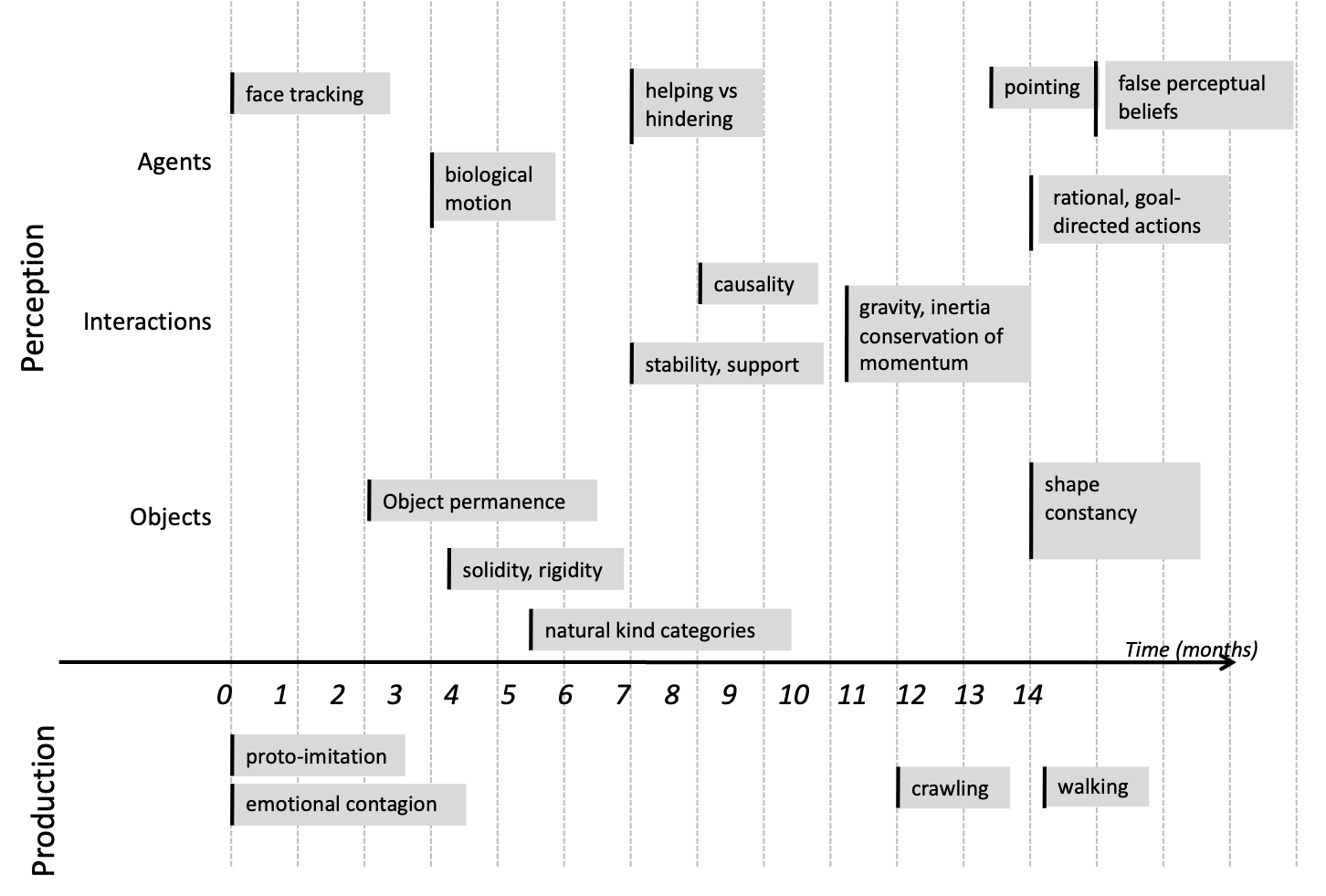}
\end{center}
   \caption{\small{Landmark of intuitive physics acquisition in infants. Each box is an experiment showing a particular ability at a given age.}}
\label{fig:timeline}
\end{figure}

Behavioral work on intuitive physics in infants and animal define a number of core conceptual components which can be tested experimentally using VEO \cite{Pauen}. Figure \ref{fig:timeline} shows a number of different landmarks in infants. 
Here, we have selected three of the most basic components and turned them into three test blocks (see \ref{tab:blocks}), each one corresponding to a core principle of intuitive physics, and each raising its particular machine vision challenge. The first two blocks are related to the conservation through time of intrinsic properties of objects. Object permanence (O1), corresponds to the fact that objects continuously exist through time and do not pop in or out of existence. This turns into the computational challenge of tracking objects through occlusion. The second block, shape constancy (O2) describes the tendency of rigid objects to preserve their shape through time. This principle is more challenging than the preceding one, because even rigid objects undergo a change in appearance due to other factors (illumination, distance, viewpoint, partial occlusion, etc.).
The final block (O3) relate to object's trajectories, and posit that each object's motion has to be continuous through space and tile (an object cannot teleport from one place to another). This principle is distinct from object permanence and requires a to incorporate smoothness constraints on the tracking of objects (even if they are not visible). Future releases of the Benchmark will continue adding progressively more complex scenarios inpired by Figure  \ref{fig:timeline}, including object interactions and agent motion.

\begin{table*}
\caption{List of the conceptual blocks of the Intuitive Physics Framework.} \label{tab:blocks}
\begin{tabular}{p{0.25\linewidth}p{0.32\linewidth}p{0.36\linewidth}}
\hline
	Block Name & Physical principles &Computational challenge  \\ \hline
	O1.\,Object permanence & \small{Objects don't pop in and out of existence} & Occlusion-resistant object tracking  \\ 
	O2.\,Shape constancy & Objects keep their shapes  & Appearance-robust object tracking    \\ 
	O3.\,\small{Spatio-temporal continuity} & Trajectories of objects are continuous  & Tracking/predicting object trajectories  \\ 
\hline
\end{tabular}
\end{table*}

\subsection{Pixels matched quadruplets}
\begin{figure}[t]
\begin{center}
   \includegraphics[width=0.9\linewidth]{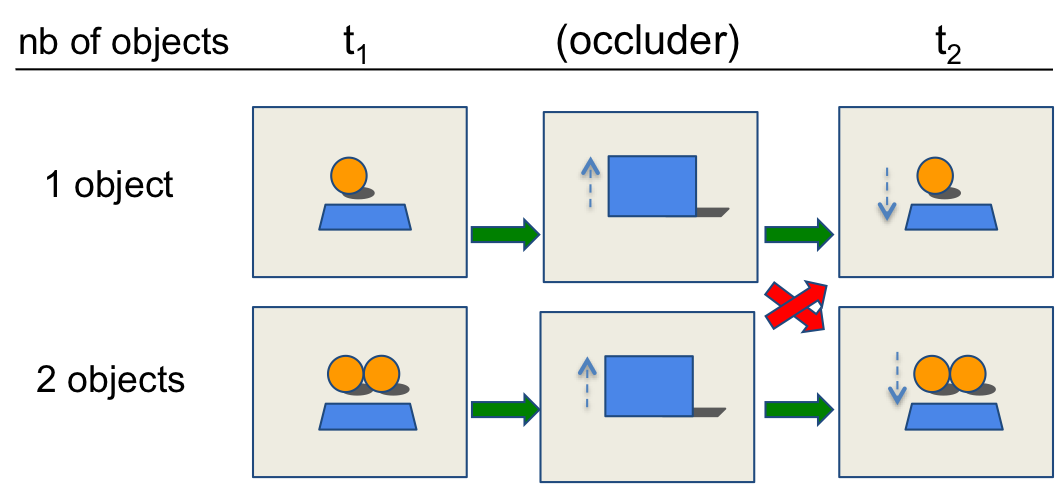}
\end{center}
   \caption{\small{Illustration of the minimal sets design with object permanence. Schematic description of a static condition with one vs. two objects and one occluder. In the two possible movies (green arrows), the number of objects remains constant despite the occlusion. In the two impossible movies (red arrows), the number of objects changes (goes from 1 to 2 or from 2 to 1).}}
\label{fig:permanence}
\end{figure}

An important design principle of our evaluation framework relates to the organization of the possible and impossible movies in extremely well matched sets to minimize the existence of low level biases. This is illustrated in Figure \ref{fig:permanence} for object permanence.
We constructed matched sets comprising four movies, which contain an initial scene at time $t_1$ (either one or two objects), and a final scene at time $t_2$ (either one or two objects), separated by a potential occlusion by a screen which is raised and then lowered for a variable amount of time. At its maximal height, the screen completely occludes the objects so that it is impossible to know, in this frame, how many objects are behind the occluder.

The four movies are constructed by combining the two possible beginnings with the two possible endings, giving rise to two possible ($1{\rightarrow}1$ and $2{\rightarrow}2$) and two impossible ($1{\rightarrow}2$ and $2{\rightarrow}1$) movies. 
Importantly, across these 4 movies, the possible and impossible ones are made of frames with the exact same pixels, 
the only factor distinguishing them being the temporal coherence of these frames. Such a design is intended to make it difficult for algorithms to use cheap tricks to distinguish possible from impossible movies by focusing on low level details, but rather requires models to focus on higher level temporal dependencies between frames. 

\subsection{Parametric task complexity}
Our second design principle is that in each block, we vary the stimulus complexity in a parametric fashion. In the case of the object permanence block, for instance, stimulus complexity can vary according to three dimensions. The first dimension is whether the change in number of objects occurs in plain view (\textit{visible}) or hidden behind an occluder (\textit{occluded}). A change in plain view is evidently easier to detect whereas a hidden change requires an element of short term memory in order to keep a trace of the object's through time. The second dimension is the complexity of the object's motion. Tracking an immobile object is easier than if the object has a complicated motion; we introduce three levels of motion complexity (static, dynamic 1, and dynamic 2). The third dimension is the number of objects involved in the scene. This tests for the attentional capacity of the system as defined by the number of objects it can track simultaneously. Manipulating stimulus complexity is important to establish the limit of what a vision system can do, and where it will fail. For instance, humans are well known to fail when the number of objects to track simultaneously is greater than four \cite{pylyshyn1988tracking}. In total, a given block contains 2 by 3 by 3, ie, 18 different scenarios varying in difficulty (see Table x).

\subsection{Procedurally generated variability}
Our final design principle is that each scenario within each block is procedurally generated in 200 examplars with random variations in objects shapes and textures, distances, trajectories, occluder motion and position of the camera. This is to minimize the possibility of focusing on only certain frames or parts of the screen to solve the task. Note that the dynamic 2 condition contains two violations instead of one. These violations are inverses of one another, such that the first and last segment of the impossible video clips are compatible with with the absence of any violation in the central part of the video (for instance, the initial and final number of objects is the same, but varies in the middle of the clip). This insures that physical violations occur in unpredictable moments in a video clip. 

\begin{figure}[h]
\begin{center}
   \includegraphics[width=0.9\linewidth]{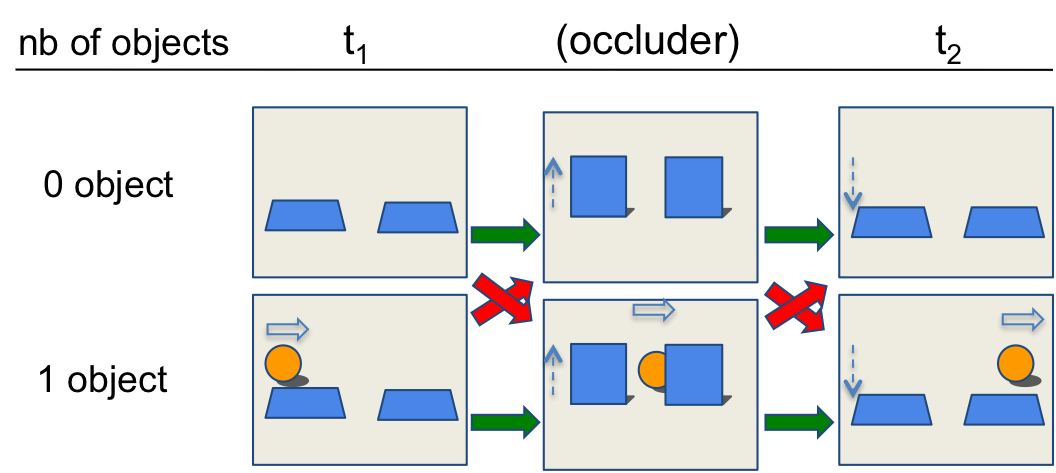}
\end{center}
   \caption{Illustration of the 'dynamic 2' condition. In the two possible movies (green arrows), the number of objects remains constant despite the occlusion. In the two impossible movies (red arrows), the number of objects changes temporarily (goes from 0 to 1 to 0 or from 1 to 0 to 1).}
\label{fig:dyna}
\end{figure}

\subsection{The possible versus impossible discrimination metric}
Our evaluation metrics depend on the system's ability to compute a \textit{plausibility score} $P(x)$ given a movie $x$. Because the test movies are structured in $N$ matched k-uplets (in Figure \ref{fig:permanence}, $k=4$) of positive and negative movies $S_{i=1..N}=\{Pos_{i}^{1}..Pos_{i}^{k},Imp_{i}^{1}..Imp_{i}^{k}\}$, we derive two different metrics. The \textit{relative} error rate $L_{R}$ computes a score within each set. It requires only that within a set, the positive movies are more plausible than the negative movies.

\begin{equation}\label{eq:L_R}
L_{R}=\frac{1}{N}\sum_{i}{\mathbbm{1}_{\sum_{j}P(Pos_{i}^{j}) < \sum_{j}P(Imp_{i}^{j})}}
\end{equation}

The absolute error rate $L_A$ requires that globally, the score of the positive movies is greater than the score of the negative movies. It is computed as:

\begin{equation}\label{eq:L_A}
L_{A}=1-AUC(\{i,j; P(Pos_{i}^{j})\}, \{i,j;  P(Imp_{i}^{j})\})
\end{equation}

Where $AUC$ is the Area Under the ROC Curve, which plots the true positive rate against the false positive rate at various threshold settings.

\subsection{Implementation}

\begin{figure}
  \begin{subfigure}{\linewidth}
  \includegraphics[width=.2\linewidth]{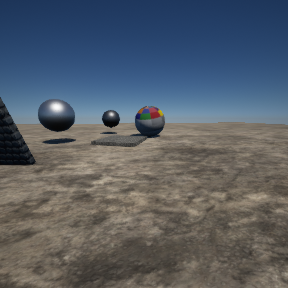}\hfill
  \includegraphics[width=.2\linewidth]{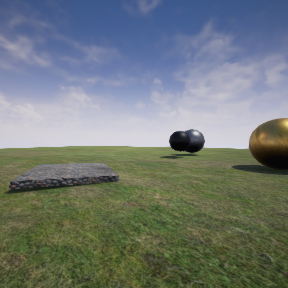}\hfill
  \includegraphics[width=.2\linewidth]{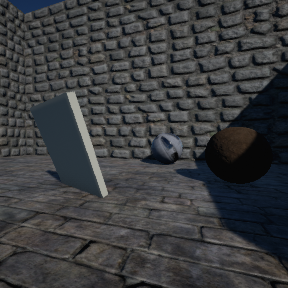}\hfill
  \includegraphics[width=.2\linewidth]{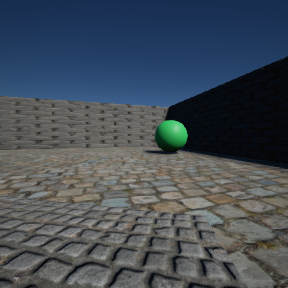}\hfill
  \includegraphics[width=.2\linewidth]{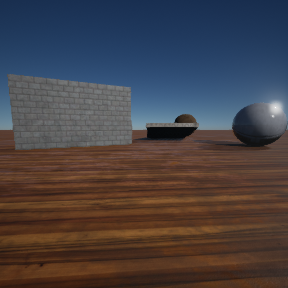}
  \end{subfigure}\par\medskip
  \caption{Examples of frames from the training set.}\label{fig:examples}
\end{figure}

The video clips in \IntPhys{} are constructed with Unreal Engine 4.0 ((UnrealEnginePython 4.19; See Figure \ref{fig:examples} for some examples). They are accessible in \url{https://intphys.com/}.

\subsubsection{The training set}

The training set contains a large variety of objects interacting one with another, occluders, textures, etc. It is composed of 15K videos of possible events (around 7 seconds each at 15fps), totalling 21 hours of videos. There are no video of impossible events, but the training set contains the objects and occluders presented in the test set. Each video is delivered as stacks of raw image (288 x 288 pixels), totalling 157Gb of uncompressed data. We also release the source code for data generation, allowing users to generate a larger training set if desired.

\subsubsection{The dev and test sets}

As described above, each of the three blocks contain 18 different scenario. In the dev set, each scenario is instantiated by 20 different renderings resulting in 360 movies per block (30 min, 3.7Gb). In the test set, a scenario has 200 different renderings of these scenarios, resulting in a total of 3600 movies per block (5h,37Gb). All of the objects and textures of the dev and test sets are present in the training set.

The purpose of the dev set released in \IntPhys{} V1.0 is to help in the selection of an appropriate plausibility score, and in the comparison of various architectures (hyper-parameters), but it should \textit{not} serve to train the model's parameters (this should be done only with the training set). This is why the dev set is kept intentionally small. The test set has more statistical power and enables a fine grained evaluation of the results across the different movie subtypes. Video examples of each blocks are available on the project page \url{www.intphys.com/}.

\subsubsection{Metadata}

Even though the spirit of \IntPhys{}  is the unsupervised learning of intuitive physics, we do provide in the test set additional information which may help the learner. The first one is the depth field for each image. This is not unreasonable, given that in infants, stereo vision and motion cues could provide an approximation of this information. The second one is object instance segmentation masks, which are helpful to recover abstract object positions but only provide local low-level information. Importantly, these masks are not linked to a specific object ID, and are randomly shuffled at each time frame. Linking instance segmentation masks to unique object IDs through time is indeed part of the object permanence problem that systems are supposed to solve. Similarly, if an object is partly occluded and appears as two pieces of object the two pieces will receive a different instance mask. 

In the train set, we do provide additional metadata about the ground truth 3D position of each object, the position of the camera, and the link between object IDs and instance masks. These metadata are not present in the dev or test sets.

\subsubsection{Submission procedure}

For each movie in the dev or test set, the model should issue a scalar plausibility score. This number together with the movie ID is then fed to the evaluation software which outputs two tables of results, one for the absolute score and the other for the relative score. 

The evaluation software is provided for the dev set, but not the test set. For evaluating on the test set, participants are invited to submit their system and results on CodaLab (see \url{www.intphys.com}) 
and their results will be registered and time-stamped on the website leaderboard.

\section{Two baseline learning models}\label{sec:baseline}

In this section, we present two learning systems which attempt to learn intuitive physics in an unsupervised/self-supervised observational setting. One can imagine an agent who only sees physical interactions between objects seen from a first-person perspective, but cannot move nor interact with them. Arguably, this is a much more empoverished learning situation than that faced by infants, who can explore and interact with their environment, even with the limited motor abilities of their first year of life. It is however interesting to establish how far one can get with such simplified inputs, which are easy to gather in abundant amounts in the real world with video cameras. In addition, this enables an easier comparison between models, because they all get the same training data. 

In a setup like this, a rich source of learning information resides in the temporal dependencies between successive frames. Based on the literature on next frame prediction, we propose two neural network models, trained on a future frame objective. Our first model has a CNN encoder-decoder structure and the second is a conditional Generative Adversarial Network (GAN, \cite{goodfellow2014generative}), with a similar structure as DCGAN \cite{DBLP:journals/corr/RadfordMC15}. For both model architectures, we investigate two different training procedures: in the first, we train models to predict short-future images with a prediction span of 5 frames; in the second, we predict long-future images with a prediction span of 35 frames.

Preliminary work with predictions at the pixel level revealed that our models failed at predicting convincing object motions, especially for small objects on a rich background. For this reason, we switched to computing predictions at a higher level, using object masks. We use the metadata provided in the benchmark training (see section \ref{sec:training}) set to train a semantic mask Deep Neural Network (DNN). This DNN uses a resnet-18 pretrained on Imagenet to extract features from the image, from which a deconvolution network is trained to predict the semantic mask (distinguishing three types of entities: background, occluders and objects). We then use this mask as input to a prediction component which predicts future masks based on past ones. 

To evaluate these models on our benchmark, our system needs to output a plausibility score for each movie. For this, we compute the prediction loss along the movie. Given past frames, a plausibility score for the frame $f_t$ can be derived by comparing $f_t$ with the prediction $\hat{f}_t$. Like in \cite{Fragkiadaki2016}, we use the analogy with an agent running an internal simulation (“visual imagination”); here we assimilate a greater distance between prediction and observation with a lower plausibility. In subsection \ref{subsec:VPS} we detail how we aggregate the scores of all frames into a plausibility score for the whole video.

\subsection{Models}

Through out the movie, our models take as input two frames $(f_{i_1}, f_{i_2})$ and predict a future frame $f_{target}$. The prediction span is independent from the model's architecture and depends only on the triplets $(f_{i_1}, f_{i_2}, f_{target})$ provided during the training phase. Our two architectures are trained either on a short term prediction task (5 frames in the future), or a long term prediction task (35 frames). Intuitively, short-term prediction will be more robust, but long-term prediction will allow the model to grasp long-term dependencies and deal with long occlusions.\\

\subsubsection{CNN encoder-decoder}
We use a resnet-18 \cite{he2016deep} pretrained on Imagenet \cite{Russakovsky2015} to extract features from input frames $(f_{i_1}, f_{i_2})$. A deconvolution network is trained to predict the semantic mask of future frame $f_{target}$ conditioned to these features, using a L2 loss. See details in Table \ref{tab:resnet_ae}\\

\begin{table}[h]
\caption{CNN for forward prediction (13941315 parameters). BN stands for batch-normalization.} \label{tab:resnet_ae}
\begin{center}
    \begin{footnotesize}
	\begin{tabular}{|c|}
	\hline
	
	Input frames \\
	2 x 3 x 64 x 64 \\
	\hline
	7 first layers of resnet-18 (pretrained, frozen weights)\\
	applied to each frame\\
	\hline
    Reshape 1 x 16384\\
    \hline
	FC 16384 $\rightarrow$ 512\\
	\hline
	FC 512 $\rightarrow$ 8192\\
	\hline
	Reshape 128 x 8 x 8\\
	\hline
	UpSamplingNearest(2), 3 x 3 Conv. 128 - 1 str., BN, ReLU\\
	\hline
	UpSamplingNearest(2), 3 x 3 Conv. 64 - 1 str., BN, ReLU\\
	\hline
	UpSamplingNearest(2), 3 x 3 Conv. 3 - 1 str., BN, ReLU\\
	\hline
	3 sigmoid\\
	Target mask\\
	\hline
	\end{tabular}
	\end{footnotesize}
\end{center}
\end{table}

\subsubsection{Generative Adversarial Network}
As a second model, we propose a conditional generative adversarial network (GAN, \cite{DBLP:journals/corr/MirzaO14}) that takes as input predicted semantic masks from frames $(f_{i_1}, f_{i_2})$, and predicts the semantic mask of future frame $f_{target}$. In this setup, the discriminator has to distinguish between a mask predicted from $f_{target}$ directly (\textit{real}), and a mask predicted from past frames $(f_{i_1}, f_{i_2})$. Like in \cite{DBLP:journals/corr/DentonGF16}, our model combines a conditional approach with a similar structure as of DCGAN \cite{DBLP:journals/corr/RadfordMC15}. At test time, we derive a plausibility score by computing the conditioned discriminator's score for every conditioned frame. This is a novel approach based on the observation that the optimal discriminator $D$ computes a score for $x$ of

\begin{equation}
D(x) = \frac{P_{data}(x)}{P_{G}(x) + P_{data}(x)}
\end{equation}

For non-physical events $\hat{x}$, $P_{data}(\hat{x})=0$; therefore, as long as $P_{G}(\hat{x})>0$, $D(\hat{x})$ should be $0$ for non-physical events, and $D(x)>0$ for physical events $x$. Note that this is a strong assumption, as there is no guarantee that the generator will ever have support at the part of the distribution corresponding to impossible videos. The generator and discriminator are detailed in Table \ref{tab:G} and \ref{tab:D}, respectively.

\begin{table}[h]

\caption{Generator G (14729347 parameters). SFConv stands for spatial full convolution and BN stands for batch-normalization.} \label{tab:G}
\begin{center}
\begin{footnotesize}
	\begin{tabular}{|c|}
	\hline
	\begin{tabular}{c | c}
	\hline
	\begin{tabular}{c}
	Input masks \\
	2 x 3 x 64 x 64\\
	\hline
	4 x 4 conv 64 - 2 str., BN, ReLU\\
	\hline
	4 x 4 conv 128 - 2 str., BN, ReLU\\
	\hline
	4 x 4 conv 256 - 2 str., BN, ReLU\\
	\hline
	4 x 4 conv 512 - 2 str., BN, ReLU\\
	\hline
    4 x 4 conv 512, BN, ReLU\\
	\end{tabular}
	&
	\begin{tabular}{c}
	\\
	\\
	\\
	Noise $\in \mathbf{R}^{100}$\\
	$\sim \text{Unif}(-1,1)$\\
	\end{tabular}\\

	\hline
	\end{tabular}\\
	stack input and noise\\
	\hline
	4 x 4 SFConv. 512 - 2 str., BN, ReLU\\
	\hline
	4 x 4 SFConv. 256 - 2 str., BN, ReLU\\
	\hline
	4 x 4 SFConv. 128 - 2 str., BN, ReLU\\
	\hline
	4 x 4 SFConv. 64 - 2 str., BN, ReLU\\
	\hline
	4 x 4 SFConv. 3 - 2 str., BN, ReLU\\
	\hline
	3 sigmoid\\
	Target mask\\
	\hline
	\end{tabular}
\end{footnotesize}
\end{center}
\end{table}

\begin{small}
\begin{table}[h]
\caption{Discriminator D (7629698 parameters). BN stands for batch-normalization.}\label{tab:D}
\begin{center}
	\begin{tabular}{|c|}
	\hline
	\begin{tabular}{c | c }
	history & input\\
	2 x 3 x 64 x 64 & 3 x 64 x 64\\
	\end{tabular}\\
	\hline
	Reshape 3 x 3 x 64 x 64\\
	\hline
	4 x 4 convolution 512 - 2 strides, BN, LeakyReLU\\
	\hline
	4 x 4 convolution 254 - 2 strides, BN, LeakyReLU\\
	\hline
	4 x 4 convolution 128 - 2 strides, BN, LeakyReLU\\
	\hline
	4 x 4 convolution 64 - 2 strides, BN, LeakyReLU\\
	\hline
	4 x 4 convolution 5 - 2 strides, BN, LeakyReLU\\
	\hline
	fully-connected layer\\
	\hline
	1 sigmoid\\
	\hline
	\end{tabular}
\end{center}
\end{table}
\end{small}

\subsubsection{Training Procedure}
We separate 10\% of the training dataset to control the overfitting of our forward predictions. All our models are trained using Adam \cite{DBLP:journals/corr/KingmaB14}. For the CNN encoder-decoder we use Adam's default parameters and stop the training after one epoch. For the GAN, we use the same hyper-parameters as in \cite{DBLP:journals/corr/RadfordMC15}: we set the generator's learning rate to $8e-4$ and discriminator's learning rate to $2e-4$. On the short-term prediction task, we train the GAN for $1$ epoch; on the long-term prediction task we train it for $5$ epochs. Learning rate decays are set to $0$ and $beta1$ is set to $0.5$ for both generator and discriminator.

The code for all our experiments is available on \url{https://github.com/rronan/IntPhys-Baselines}.

\subsection{Video Plausibility Score}\label{subsec:VPS}

From forward models presented above, we can compute a plausibility score for every frame $f_{target}$, conditioned to previous frames $(f_{i_1}, f_{i_2})$. However, because the temporal positions of impossible events are not given, we must decide of a score for a video, given the scores of all its conditioned frames. An impossible event can be characterized by the presence of one or more impossible frame(s), conditioned to previous frames. Hence, a natural approach to compute a video plausibility score is to take the minimum of all conditioned frames' scores:

\begin{equation}
\text{Plaus}(v) = \min\limits_{(f_{i_1}, f_{i_2}, f_{target}) \in v}\text{Plaus}(f_{target} | f_{i_1}, f_{i_2})
\end{equation}

where $v$ is the video, and $(f_{i_1}, f_{i_2}, f_{target})$ are all the frame triplets in $v$, as given in the training phase.\\

\subsection{Results}\label{ssec:Results}

\subsubsection{Block O1}
\paragraph*{Short-term prediction} The first training procedure is a short-term prediction task; it takes as input frames $f_{t-2}, f_{t}$ and predicts $f_{t+5}$, which we note $(f_{t-2}, f_{t}) \rightarrow f_{t+5}$ in the following. We train the two architectures presented above on short-term prediction task and evaluate them on the test set. For the relative classification task, CNN encoder-decoder has an error rate of 0.09 when impossible events are visible and 0.49 when they are occluded. The GAN has an error rate of 0.15 when visible and 0.48 when occluded. For the absolute classification task, CNN encoder-decoder has a $L_A$ (see eq. \ref{eq:L_A}) of 0.33 when impossible events are visible and 0.50 when they are occluded. The GAN has a $L_A$ of 0.38 when visible and 0.50 when occluded. Results are detailed in Supplementary Materials (Tables 1, 2, 3, 4). \\
We observe that our short-term prediction models show good performances when the impossible events are visible, especially on the relative classifications task. However they perform poorly when the impossible events are occluded. This is easily explained by the fact that they have a prediction span of 5 frames, which is usually lower than the occlusion time. Hence, these models don't have enough "memory" to catch occluded impossible events.\\

\paragraph*{Long-term prediction} The second training procedure consists in a long-term prediction task: $(f_{t-5}, f_{t}) \rightarrow f_{t+35}$. For the relative classification task, CNN encoder-decoder has an error rate of 0.07 when impossible events are visible and 0.52 when they are occluded. The GAN has an error rate of 0.17 when visible and 0.48 when occluded. For the absolute classification task, CNN encoder-decoder has a $L_A$ of 0.37 when impossible events are visible and 0.50 when they are occluded. The GAN has a $L_A$ of 0.40 when visible and 0.50 when occluded. Results are detailed in Supplementary Materials (Tables 5, 6, 7, 8).
As expected, long-term models perform better than short-term models on occluded impossible events. Moreover, results on absolute classification task confirm that it is way more challenging than the relative classification task. Because some movies are more complex than others, the average score of each quadruplet of movies may vary a lot. It results in cases where one model returns a higher plausibility score to an impossible movie $M_{\{\text{imp, easy}\}}$ from an easy quadruplet than to a possible movie $M_\text{\{pos, complex\}}$ from a complex quadruplet.\\

\begin{figure*}[h]
\centering
    \begin{subfigure}[b]{0.45\textwidth}
        \includegraphics[width=\textwidth]{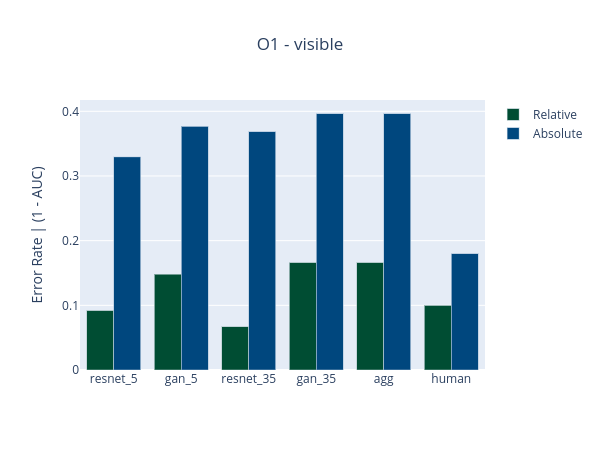}
    \end{subfigure}
    \begin{subfigure}[b]{0.45\textwidth}
        \includegraphics[width=\textwidth]{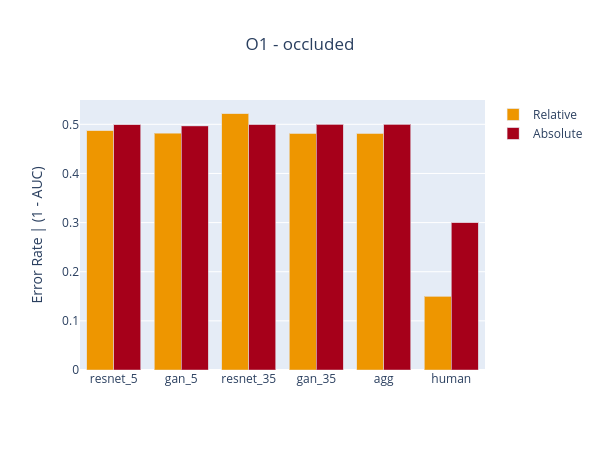}
    \end{subfigure}
   \caption{Results of our baselines on block O1, in cases where the impossible event occurs in the open (\textit{visible}) or behind an occluder (\textit{occluded}). Y-axis represents the losses $L_R$ (see Equation 1) for the relative performance and $L_A$ (see Equation 2) for the absolute performance.}
\label{fig:bar_O1}
\end{figure*}

\paragraph*{Aggregated model} On the relative classification task, the aggregated CNN encoder-decoder has an error rate of 0.07 when impossible events are visible and 0.52 when they are occluded. For the absolute classification task, CNN encoder-decoder has a $L_A$ of 0.37 when impossible events are visible and 0.50 when they are occluded. Results are detailed in Figures \ref{fig:bar_O1}, \ref{fig:bar_O2}, \ref{fig:bar_O3} and Supplementary Materials (Tables 9, 10).

\subsubsection{Block O2}
\paragraph*{Short-term prediction} For the first training procedure $(f_{t-2}, f_{t}) \rightarrow f_{t+5}$: CNN encoder-decoder has an relactive classification error rate of 0.16 when impossible events are visible and 0.49 when they are occluded. The GAN has an error rate of 0.30 when visible and 0.52 when occluded. For the absolute classification task, CNN encoder-decoder has a $L_A$ of 0.40 when impossible events are visible and 0.50 when they are occluded. The GAN has a $L_A$ of 0.43 when visible and 0.50 when occluded. Results are detailed in Supplementary Materials (Tables 11, 12, 13, 14).\\

\paragraph*{Long-term prediction} For the second training procedure $(f_{t-5}, f_{t}) \rightarrow f_{t+35}$: the CNN encoder-decoder has an error rate of 0.11 when impossible events are visible and 0.52 when they are occluded. The GAN has an error rate of 0.31 when visible and 0.50 when occluded. For the absolute classification task, CNN encoder-decoder has a $L_A$ of 0.43 when impossible events are visible and 0.50 when they are occluded. The GAN has a $L_A$ of 0.33 when visible and 0.50 when occluded. Results are detailed in Supplementary Materials (Tables 15, 16, 17, 18).\\

\begin{figure*}[h]
\centering
    \begin{subfigure}[b]{0.45\textwidth}
        \includegraphics[width=\textwidth]{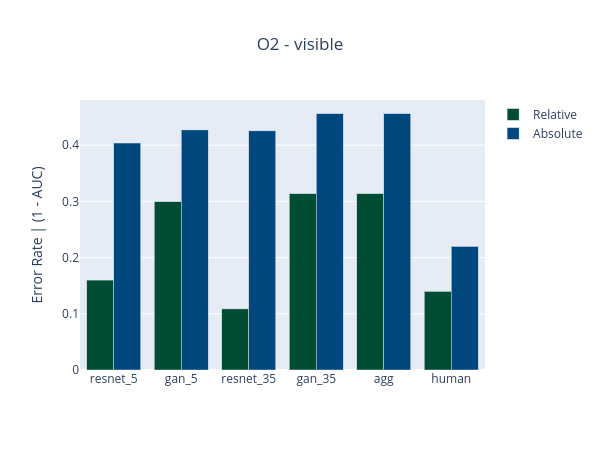}
    \end{subfigure}
    \begin{subfigure}[b]{0.45\textwidth}
        \includegraphics[width=\textwidth]{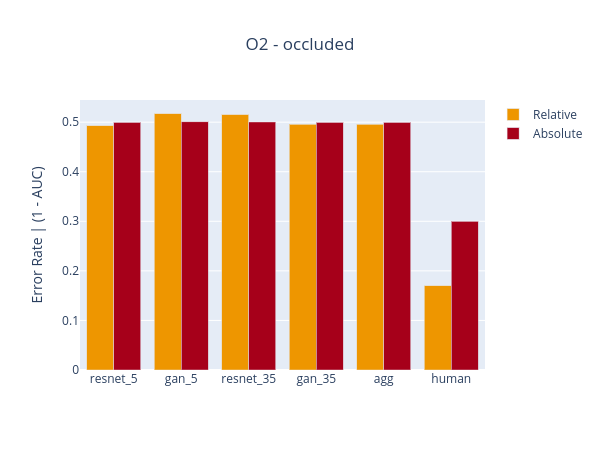}
    \end{subfigure}
   \caption{Results of our baselines on block O2, in cases where the impossible event occurs in the open (\textit{visible}) or behind an occluder (\textit{occluded}). Y-axis represents the losses $L_R$ (see Equation 1) for the relative performance and $L_A$ (see Equation 2) for the absolute performance.}
\label{fig:bar_O2}
\end{figure*}

\paragraph*{Aggregated model} On the relative classification task, the aggregated CNN encoder-decoder has an error rate of 0.11 when impossible events are visible and 0.52 when they are occluded. For the absolute classification task, CNN encoder-decoder has a $L_A$ of 0.43 when impossible events are visible and 0.50 when they are occluded. Results are detailed in Figures \ref{fig:bar_O1}, \ref{fig:bar_O2}, \ref{fig:bar_O3} and Supplementary Materials (Tables 19, 20).

\subsubsection{Block O3}
\paragraph*{Short-term prediction} For the first training procedure $(f_{t-2}, f_{t}) \rightarrow f_{t+5}$: CNN encoder-decoder has an relactive classification error rate of 0.28 when impossible events are visible and 0.49 when they are occluded. The GAN has an error rate of 0.26 when visible and 0.48 when occluded. For the absolute classification task, CNN encoder-decoder has a $L_A$ (see eq. \ref{eq:L_A}) of 0.40 when impossible events are visible and 0.50 when they are occluded. The GAN has a $L_A$ of 0.42 when visible and 0.50 when occluded. Results are detailed in Supplementary Materials (Tables 21, 22, 23, 24).\\

\paragraph*{Long-term prediction} For the second training procedure $(f_{t-5}, f_{t}) \rightarrow f_{t+35}$: the CNN encoder-decoder has an error rate of 0.32 when impossible events are visible and 0.51 when they are occluded. The GAN has an error rate of 0.34 when visible and 0.52 when occluded. For the absolute classification task, CNN encoder-decoder has a $L_A$ of 0.46 when impossible events are visible and 0.50 when they are occluded. The GAN has a $L_A$ of 0.44 when visible and 0.50 when occluded. Results are detailed in Supplementary Materials (Tables 25, 26, 27, 28).\\

\begin{figure*}[h]
\centering
    \begin{subfigure}[b]{0.45\textwidth}
        \includegraphics[width=\textwidth]{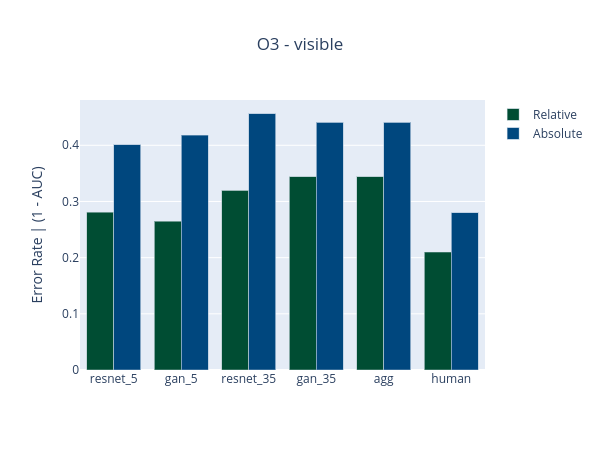}
    \end{subfigure}
    \begin{subfigure}[b]{0.45\textwidth}
        \includegraphics[width=\textwidth]{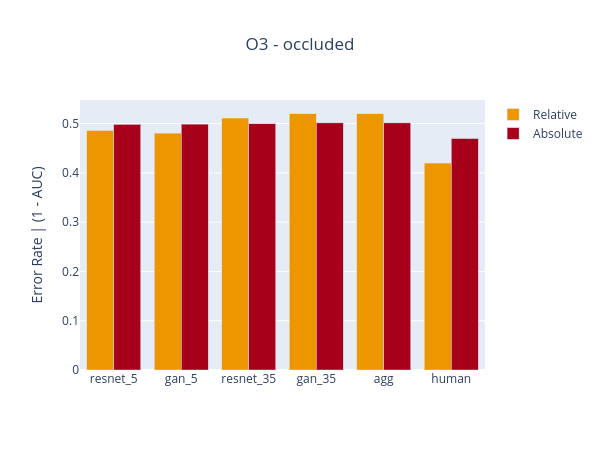}
    \end{subfigure}
   \caption{Results of our baselines on block O3, in cases where the impossible event occurs in the open (\textit{visible}) or behind an occluder (\textit{occluded}). Y-axis represents the losses $L_R$ (see Equation 1) for the relative performance and $L_A$ (see Equation 2) for the absolute performance.}
\label{fig:bar_O3}
\end{figure*}

\paragraph*{Aggregated model} On the relative classification task, the aggregated CNN encoder-decoder has an error rate of 0.32 when impossible events are visible and 0.51 when they are occluded. For the absolute classification task, CNN encoder-decoder has a $L_A$ of 0.46 when impossible events are visible and 0.50 when they are occluded. Results are detailed in Figures \ref{fig:bar_O1}, \ref{fig:bar_O2}, \ref{fig:bar_O3} and Supplementary Materials (Tables 29, 30).

As expected, we observe that models' performance decrease when impossible events are occluded. This enlightens the difficulty to perform long-term predictions in videos. We also observe that their performances vary with the types of impossible events tested. Results are the highest when testing presence / absence of object, and the lowest when testing the temporal continuity of trajectories.

\section{Human Judgements Experiment}\label{sec:human}

To give a second reference to evaluate physical understanding in models, and provide a good description of human performance on this benchmark, we presented the 3600 videos from each block to human participants using Amazon Mechanical Turk. Participants were first presented 8 examples of possible scenes from the training set, some simple, some more complex. They were told that some of the test movies were incorrect or corrupted, in that they showed events that could not possibly take place in the real world (without specifying how). Participants were each presented with 40 randomly selected videos, and were asked to score them from 1 (most implausible) to 6 (most plausible). They completed the task in about 7 minutes, and were paid \$1. A total of 540 persons participated, such that every video tested was seen by 2 different participants. A mock sample of the AMT test is available on \url{http://129.199.81.135/naive_physics_experiment/}.

\begin{table*}[h]
\centering
\caption{Average error rate on plausibility judgments collected in humans using MTurk for the \IntPhys (Block O1) test set.}
\label{tab:human_abs_O1}
\begin{tabular}{lccccccccc}
\toprule
& \multicolumn{3}{c}{Visible}& & &\multicolumn{3}{c}{Occluded}& \\ 

\cmidrule{2-4} \cmidrule{7-9}
Type of scene           &1 obj.&2 obj.&3 obj.& Total  &&1 obj. &2 obj.& 3 obj. & Total \\ 

\midrule
Static                  & 0.13 & 0.14 & 0.09 & 0.12  && 0.32  & 0.34 &   0.28 & 0.31 \\ 

Dynamic (1 violation)   & 0.15 & 0.29 & 0.27 & 0.24  && 0.24  & 0.30 &   0.33 & 0.29   \\ 

Dynamic (2 violations)  & 0.14 & 0.20 & 0.23 & 0.19  && 0.28  & 0.26 &   0.36 & 0.30  \\ 

\midrule
Total                   & 0.14 & 0.21 & 0.20 & 0.18  && 0.28  & 0.30 &   0.32 & 0.30 \\ 

\bottomrule
\end{tabular}
\end{table*}

\begin{table*}[h]
\centering
\caption{Average error rate on plausibility judgments collected in humans using MTurk for the \IntPhys (Block O2) test set.}
\label{tab:human_abs_O2}
\begin{tabular}{lccccccccc}
\toprule
& \multicolumn{3}{c}{Visible}& & &\multicolumn{3}{c}{Occluded}& \\ 

\cmidrule{2-4} \cmidrule{7-9}
Type of scene           &1 obj.&2 obj.&3 obj.& Total  &&1 obj. &2 obj.& 3 obj. & Total   \\ 

\midrule
Static                  & 0.13 & 0.18 & 0.15 & 0.16  && 0.22  & 0.33 &   0.28 & 0.28  \\ 

Dynamic (1 violation)   & 0.29 & 0.24 & 0.27 & 0.27  && 0.29  & 0.35 &   0.29 & 0.31    \\ 

Dynamic (2 violations)  & 0.21 & 0.27 & 0.26 & 0.24  && 0.32  & 0.32 &   0.29 & 0.31   \\ 

\midrule
Total                   & 0.21 & 0.23 & 0.23 & 0.22  && 0.28  & 0.33 &   0.29 & 0.30   \\ 

\bottomrule
\end{tabular}
\end{table*}

\begin{table*}[h]
\centering
\caption{Average error rate on plausibility judgments collected in humans using MTurk for the \IntPhys (Block O3) test set.}
\label{tab:human_abs_O3}
\begin{tabular}{lccccccccc}
\toprule
& \multicolumn{3}{c}{Visible}& & &\multicolumn{3}{c}{Occluded}& \\ 

\cmidrule{2-4} \cmidrule{7-9}
Type of scene           &1 obj.&2 obj.&3 obj.& Total  &&1 obj. &2 obj.& 3 obj. & Total \\

\midrule
Static                  & 0.29 & 0.32 & 0.27 & 0.29  && 0.36  & 0.36 &   0.45 & 0.39 \\

Dynamic (1 violation)   & 0.28 & 0.33 & 0.30 & 0.30  && 0.49  & 0.55 &   0.49 & 0.51 \\

Dynamic (2 violations)  & 0.23 & 0.23 & 0.26 & 0.24  && 0.47  & 0.53 &   0.55 & 0.52 \\

\midrule
Total                   & 0.27 & 0.29 & 0.28 & 0.28  && 0.44  & 0.48 &   0.50 & 0.47 \\

\bottomrule
\end{tabular}
\end{table*}

The average error rates were computed across condition, number of objects and visibility and are shown in Tables \ref{tab:human_abs_O1}, \ref{tab:human_abs_O2}, \ref{tab:human_abs_O3}. In general, observers missed violations more often when the scene was occluded; we observe error rates going from $18\%$ (visible) to $30\%$ (occluded) for block $O1$, from $22\%$ (visible) to $30\%$ (occluded) for block $O2$, from $28\%$ (visible) to $47\%$ (occluded) for block $O3$. An interesting result is that the score of humans on block $O3$ is close to chance when objects are visible. This shows that humans have trouble to detect changes in velocity of objects, when these changes occur when the object is occluded. We also observe an increase in error going from static to dynamic 1 (one occlusion) and from dynamic 1 to dynamic 2 (two occlusions), but this pattern was only consistently observed in the occluded condition. For visible scenario, the dynamic 1 appeared more difficult than the dynamic 2. This was probably due to the fact that when objects are visible, the dynamic 2 impossible scenarios contain two local discontinuities and are therefore easier to spot than when one discontinuity only is present. When the discontinuities occurred behind the occluder, the pattern of difficulties was reversed, presumably because participants started using heuristics, such as checking that the number of objects at the beginning is the same as at the end, and therefore missed the intermediate disappearance of an object.  

These results suggest that human participants are not responding according to the gold standard laws of physics due to limitations in attentional capacity - and this, even though the number of objects to track is below the theoretical limit of 4 objects. The performance of human observers can thus serve as a reference besides ground truth, especially for systems intended to model human perception.

Interestingly, we observe similar patterns of performance between models and humans (see Figures \ref{fig:bar_O1}, \ref{fig:bar_O2}, \ref{fig:bar_O3}), with increasing error rates from blocks $O1$ to $O3$. As expected, both humans and models show higher error rates when the considered impossible event is occluded.  

\section{Related work}\label{sec:related}

The modeling of intuitive physics has been addressed mostly through systems trained with some form of future prediction as a training objective. Some studies have investigated models for predicting the stability and forward modeling the dynamics of towers of blocks (\cite{BattagliaHT13,LererGF16,Zhang2016,Li2016,Mirza2017,li2016visual}). \cite{BattagliaHT13} proposes a model based on an intuitive physics engine, \cite{LererGF16} and \cite{Li2016} follow a supervised approach using Convolutional Neural Networks (CNNs), \cite{Zhang2016} makes a comparison between simulation-based models and CNN-based models, \cite{Mirza2017} improves the predictions of a CNN model by providing it with a prediction of a generative model. In \cite{mathieu2015deep}, authors propose different feature learning strategies (multi-scale architecture, adversarial training method, image gradient difference loss function) to predict future frames in raw videos. 

Other models use more structured representation of objects to derive longer-term predictions. In \cite{NIPS2016_6418} and \cite{chang2016compositional}, authors learn objects dynamics by modelling their pairwise interactions and predicting the resulting objects states representation (e.g. position / velocity / object intrinsic properties) .
In \cite{2017arXiv170601433W}, \cite{Fraccaro2017} and \cite{ehrhardt2017learning} authors combine factored latent object representations, object centric dynamic models and visual encoders. Each frame is parsed into a set of object state representations, which are used as input of a dynamic model. In \cite{Fraccaro2017} and \cite{ehrhardt2017learning}, authors use a visual decoder to reconstruct the future frames, allowing the model to learn from raw (though synthetic) videos.

Regarding evaluation and benchmarks, apart from frame prediction datasets, which are not strictly speaking about intuitive physics, one can distinguish the Visual Newtonian Dynamics (VIND) dataset
which includes more than 6000 videos with bounding boxes on key objects across frames, and annotated with a 3D plane which would most closely fit the object trajectory \cite{MottaghiRGF16}. There is also recent dataset proposed by a DeepMind team \cite{DBLP:journals/corr/abs-1804-01128}. This last dataset seems very similar to ours. It is also inspired by the developmental literature and based on the violation of expectation principles and is structured around 3 blocks similar to our first 3 blocks (object permanence, shape constancy, continuity) and two other ones (solidity and containment). The number and characteristics of this dataset is not known at present. From the sample videos, two differences emerged: our dataset is better matched in terms of quadruplets of clips controlled at the level of the pixels, and our dataset has a factorial manipulation of scene and movement complexity. It would be interesting to explore the possibility to merge these two datasets, as well as add more blocks in order to increase the diversity and coverage of the physical phenomena.

\section{Discussion}\label{sec:conc}
We presented  \IntPhys{}, a benchmark for measuring intuitive physics in artificial vision systems inspired by research on conceptual development in infants. To pass the benchmark, a system is asked to return a plausibility score for each video clip. The system's performance is assessed by measuring its ability to discriminate possible from impossible videos illustrating several types of physical principles.
Naive humans were tested on the same dataset, to give an idea of what performance could be expected by a good model. These results show error rates increasing with the presence of occlusion, but not with number of objects. This is congruent with data showing that humans can track up to three objects simultaneously. We presented two unsupervised learning models based on semantic masks, which learn from a training set only composed of physically plausible clips, and are tested on the same block as the humans.  

The computational system generally performed poorly compared to humans but obtained above chance performance using a mask prediction task, with a very strong effect of the presence of occlusion. The relative success of the semantic mask prediction system compared to what we originally found with pixel-based systems indicates that operating at a more abstract level is a worthwhile pursuing strategy when it comes to modeling intuitive physics. Future work will explore alternative ways of constructing this abstract representation in particular instance masks and object detection bounding boxes. In addition, enriching the training through embedding the learner in an interactive version of the environment could add more information for the learning of the physics of macroscopic objects.

In brief, the systematic way of constructing the \IntPhys{} Benchmark shows that it is possible to adapt developmental paradigm in a machine learning setting, and that the resulting benchmark is a relatively challenging one. The three blocks that we present here could be extended to cover more aspects of object perception, including more difficult ones like interactions between objects, or prediction of trajectories of animated agents. As we discussed in the introduction, this benchmark only provides unit tests regarding the computation of prediction probabilities of object positions based on past frames. Further work will be needed to construct benchmarks testing how theses probabilities can be used by a system to make decision or plan trajectories. 

\newpage
\bibliographystyle{IEEEtran}
\bibliography{egbib.bib}

\newpage
\appendixpage

\section{Model results (detailed)}

\begin{small}

\begin{table*}[h]
\centering
\caption{ Block O1 | Model: CNN (short-term prediction task) | Relative classification ($L_R$)}
\label{tab:modelRel}
\begin{tabular}{lccccccccc}
\toprule
& \multicolumn{3}{c}{Visible}& & &\multicolumn{3}{c}{Occluded}& \\ 

\cmidrule{2-4} \cmidrule{7-9}
Type of scene           &1 obj.&2 obj.&3 obj.& Total  &&1 obj. &2 obj.& 3 obj. & Total \\

\midrule
Static                  & 0.00 & 0.00 & 0.00 & 0.00  && 0.49  & 0.52 &   0.41 & 0.47 \\

Dynamic (1 violation)   & 0.00 & 0.22 & 0.27 & 0.17  && 0.51  & 0.47 &   0.49 & 0.49 \\

Dynamic (2 violations)  & 0.00 & 0.13 & 0.20 & 0.11  && 0.50  & 0.50 &   0.49 & 0.50 \\

\midrule
Total                   & 0.00 & 0.12 & 0.16 & 0.09  && 0.50  & 0.50 &   0.46 & 0.49 \\

\bottomrule
\end{tabular}
\end{table*}

\begin{table*}[h]
\centering
\caption{ Block O1 | Model: CNN (short-term prediction task) | Absolute classification ($L_A$)}
\label{tab:modelRel}
\begin{tabular}{lccccccccc}
\toprule
& \multicolumn{3}{c}{Visible}& & &\multicolumn{3}{c}{Occluded}& \\ 

\cmidrule{2-4} \cmidrule{7-9}
Type of scene           &1 obj.&2 obj.&3 obj.& Total  &&1 obj. &2 obj.& 3 obj. & Total \\

\midrule
Static                  & 0.15 & 0.17 & 0.19 & 0.17  && 0.50  & 0.50 &   0.49 & 0.50 \\

Dynamic (1 violation)   & 0.32 & 0.44 & 0.47 & 0.41  && 0.50  & 0.50 &   0.50 & 0.50 \\

Dynamic (2 violations)  & 0.33 & 0.43 & 0.47 & 0.41  && 0.50  & 0.50 &   0.50 & 0.50 \\

\midrule
Total                   & 0.26 & 0.35 & 0.38 & 0.33  && 0.50  & 0.50 &   0.50 & 0.50 \\

\bottomrule
\end{tabular}
\end{table*}

\begin{table*}[h]
\centering
\caption{ Block O1 | Model: GAN (short-term prediction task) | Relative classification ($L_R$)}
\label{tab:modelRel}
\begin{tabular}{lccccccccc}
\toprule
& \multicolumn{3}{c}{Visible}& & &\multicolumn{3}{c}{Occluded}& \\ 

\cmidrule{2-4} \cmidrule{7-9}
Type of scene           &1 obj.&2 obj.&3 obj.& Total  &&1 obj. &2 obj.& 3 obj. & Total \\

\midrule
Static                  & 0.00 & 0.00 & 0.00 & 0.00  && 0.44  & 0.45 &   0.53 & 0.48 \\

Dynamic (1 violation)   & 0.00 & 0.35 & 0.39 & 0.25  && 0.44  & 0.50 &   0.47 & 0.47 \\

Dynamic (2 violations)  & 0.00 & 0.21 & 0.39 & 0.20  && 0.51  & 0.50 &   0.49 & 0.50 \\

\midrule
Total                   & 0.00 & 0.18 & 0.26 & 0.15  && 0.46  & 0.48 &   0.50 & 0.48 \\

\bottomrule
\end{tabular}
\end{table*}

\begin{table*}[h]
\centering
\caption{ Block O1 | Model: GAN (short-term prediction task) | Absolute classification ($L_A$)}
\label{tab:modelRel}
\begin{tabular}{lccccccccc}
\toprule
& \multicolumn{3}{c}{Visible}& & &\multicolumn{3}{c}{Occluded}& \\ 

\cmidrule{2-4} \cmidrule{7-9}
Type of scene           &1 obj.&2 obj.&3 obj.& Total  &&1 obj. &2 obj.& 3 obj. & Total \\

\midrule
Static                  & 0.23 & 0.31 & 0.32 & 0.28  && 0.50  & 0.50 &   0.50 & 0.50 \\

Dynamic (1 violation)   & 0.33 & 0.47 & 0.50 & 0.43  && 0.49  & 0.49 &   0.50 & 0.49 \\

Dynamic (2 violations)  & 0.34 & 0.44 & 0.46 & 0.41  && 0.50  & 0.50 &   0.50 & 0.50 \\

\midrule
Total                   & 0.30 & 0.41 & 0.43 & 0.38  && 0.49  & 0.50 &   0.50 & 0.50 \\

\bottomrule
\end{tabular}
\end{table*}

\begin{table*}[h]
\centering
\caption{ Block O1 | Model: CNN (long-term prediction task) | Relative classification ($L_R$)}
\label{tab:modelRel}
\begin{tabular}{lccccccccc}
\toprule
& \multicolumn{3}{c}{Visible}& & &\multicolumn{3}{c}{Occluded}& \\ 

\cmidrule{2-4} \cmidrule{7-9}
Type of scene           &1 obj.&2 obj.&3 obj.& Total  &&1 obj. &2 obj.& 3 obj. & Total \\

\midrule
Static                  & 0.00 & 0.00 & 0.00 & 0.00  && 0.52  & 0.55 &   0.51 & 0.53 \\

Dynamic (1 violation)   & 0.00 & 0.13 & 0.22 & 0.12  && 0.49  & 0.53 &   0.48 & 0.50 \\

Dynamic (2 violations)  & 0.00 & 0.06 & 0.20 & 0.09  && 0.53  & 0.48 &   0.60 & 0.54 \\

\midrule
Total                   & 0.00 & 0.06 & 0.14 & 0.07  && 0.51  & 0.52 &   0.53 & 0.52 \\

\bottomrule
\end{tabular}
\end{table*}

\begin{table*}[h]
\centering
\caption{ Block O1 | Model: CNN (long-term prediction task) | Absolute classification ($L_A$)}
\label{tab:modelRel}
\begin{tabular}{lccccccccc}
\toprule
& \multicolumn{3}{c}{Visible}& & &\multicolumn{3}{c}{Occluded}& \\ 

\cmidrule{2-4} \cmidrule{7-9}
Type of scene           &1 obj.&2 obj.&3 obj.& Total  &&1 obj. &2 obj.& 3 obj. & Total \\

\midrule
Static                  & 0.30 & 0.34 & 0.36 & 0.33  && 0.50  & 0.50 &   0.50 & 0.50 \\

Dynamic (1 violation)   & 0.30 & 0.43 & 0.44 & 0.39  && 0.50  & 0.50 &   0.50 & 0.50 \\

Dynamic (2 violations)  & 0.32 & 0.40 & 0.43 & 0.39  && 0.50  & 0.50 &   0.50 & 0.50 \\

\midrule
Total                   & 0.31 & 0.39 & 0.41 & 0.37  && 0.50  & 0.50 &   0.50 & 0.50 \\

\bottomrule
\end{tabular}
\end{table*}

\begin{table*}[h]
\centering
\caption{ Block O1 | Model: GAN (long-term prediction task) | Relative classification ($L_R$)}
\label{tab:modelRel}
\begin{tabular}{lccccccccc}
\toprule
& \multicolumn{3}{c}{Visible}& & &\multicolumn{3}{c}{Occluded}& \\ 

\cmidrule{2-4} \cmidrule{7-9}
Type of scene           &1 obj.&2 obj.&3 obj.& Total  &&1 obj. &2 obj.& 3 obj. & Total \\

\midrule
Static                  & 0.00 & 0.01 & 0.00 & 0.00  && 0.41  & 0.58 &   0.57 & 0.52 \\

Dynamic (1 violation)   & 0.00 & 0.28 & 0.45 & 0.24  && 0.39  & 0.56 &   0.54 & 0.50 \\

Dynamic (2 violations)  & 0.01 & 0.29 & 0.46 & 0.25  && 0.43  & 0.46 &   0.40 & 0.43 \\

\midrule
Total                   & 0.00 & 0.19 & 0.30 & 0.17  && 0.41  & 0.54 &   0.50 & 0.48 \\

\bottomrule
\end{tabular}
\end{table*}

\begin{table*}[h]
\centering
\caption{ Block O1 | Model: GAN (long-term prediction task) | Absolute classification ($L_A$)}
\label{tab:modelRel}
\begin{tabular}{lccccccccc}
\toprule
& \multicolumn{3}{c}{Visible}& & &\multicolumn{3}{c}{Occluded}& \\ 

\cmidrule{2-4} \cmidrule{7-9}
Type of scene           &1 obj.&2 obj.&3 obj.& Total  &&1 obj. &2 obj.& 3 obj. & Total \\

\midrule
Static                  & 0.26 & 0.33 & 0.37 & 0.32  && 0.50  & 0.50 &   0.50 & 0.50 \\

Dynamic (1 violation)   & 0.36 & 0.46 & 0.49 & 0.44  && 0.49  & 0.50 &   0.50 & 0.50 \\

Dynamic (2 violations)  & 0.35 & 0.47 & 0.48 & 0.43  && 0.50  & 0.50 &   0.50 & 0.50 \\

\midrule
Total                   & 0.32 & 0.42 & 0.45 & 0.40  && 0.50  & 0.50 &   0.50 & 0.50 \\

\bottomrule
\end{tabular}
\end{table*}

\begin{table*}[h]
\centering
\caption{ Block O1 | Model: CNN aggregated | Relative classification ($L_R$)}
\label{tab:modelRel}
\begin{tabular}{lccccccccc}
\toprule
& \multicolumn{3}{c}{Visible}& & &\multicolumn{3}{c}{Occluded}& \\ 

\cmidrule{2-4} \cmidrule{7-9}
Type of scene           &1 obj.&2 obj.&3 obj.& Total  &&1 obj. &2 obj.& 3 obj. & Total \\

\midrule
Static                  & 0.00 & 0.00 & 0.00 & 0.00  && 0.52  & 0.55 &   0.51 & 0.53 \\

Dynamic (1 violation)   & 0.00 & 0.13 & 0.22 & 0.12  && 0.49  & 0.53 &   0.48 & 0.50 \\

Dynamic (2 violations)  & 0.00 & 0.06 & 0.20 & 0.09  && 0.53  & 0.48 &   0.60 & 0.54 \\

\midrule
Total                   & 0.00 & 0.06 & 0.14 & 0.07  && 0.51  & 0.52 &   0.53 & 0.52 \\

\bottomrule
\end{tabular}
\end{table*}

\begin{table*}[h]
\centering
\caption{ Block O1 | Model: CNN aggregated | Absolute classification ($L_A$)}
\label{tab:modelRel}
\begin{tabular}{lccccccccc}
\toprule
& \multicolumn{3}{c}{Visible}& & &\multicolumn{3}{c}{Occluded}& \\ 

\cmidrule{2-4} \cmidrule{7-9}
Type of scene           &1 obj.&2 obj.&3 obj.& Total  &&1 obj. &2 obj.& 3 obj. & Total \\

\midrule
Static                  & 0.30 & 0.34 & 0.36 & 0.33  && 0.50  & 0.50 &   0.50 & 0.50 \\

Dynamic (1 violation)   & 0.30 & 0.43 & 0.44 & 0.39  && 0.50  & 0.50 &   0.50 & 0.50 \\

Dynamic (2 violations)  & 0.32 & 0.40 & 0.43 & 0.39  && 0.50  & 0.50 &   0.50 & 0.50 \\

\midrule
Total                   & 0.31 & 0.39 & 0.41 & 0.37  && 0.50  & 0.50 &   0.50 & 0.50 \\

\bottomrule
\end{tabular}
\end{table*}

\begin{table*}[h]
\centering
\caption{ Block O2 | Model: CNN (short-term prediction task) | Relative classification ($L_R$)}
\label{tab:modelRel}
\begin{tabular}{lccccccccc}
\toprule
& \multicolumn{3}{c}{Visible}& & &\multicolumn{3}{c}{Occluded}& \\ 

\cmidrule{2-4} \cmidrule{7-9}
Type of scene           &1 obj.&2 obj.&3 obj.& Total  &&1 obj. &2 obj.& 3 obj. & Total \\

\midrule
Static                  & 0.00 & 0.00 & 0.00 & 0.00  && 0.48  & 0.49 &   0.47 & 0.48 \\

Dynamic (1 violation)   & 0.18 & 0.26 & 0.40 & 0.28  && 0.50  & 0.49 &   0.50 & 0.50 \\

Dynamic (2 violations)  & 0.12 & 0.16 & 0.32 & 0.20  && 0.50  & 0.50 &   0.50 & 0.50 \\

\midrule
Total                   & 0.10 & 0.14 & 0.24 & 0.16  && 0.49  & 0.49 &   0.49 & 0.49 \\

\bottomrule
\end{tabular}
\end{table*}

\begin{table*}[h]
\centering
\caption{ Block O2 | Model: CNN (short-term prediction task) | Absolute classification ($L_A$)}
\label{tab:modelRel}
\begin{tabular}{lccccccccc}
\toprule
& \multicolumn{3}{c}{Visible}& & &\multicolumn{3}{c}{Occluded}& \\ 

\cmidrule{2-4} \cmidrule{7-9}
Type of scene           &1 obj.&2 obj.&3 obj.& Total  &&1 obj. &2 obj.& 3 obj. & Total \\

\midrule
Static                  & 0.22 & 0.29 & 0.28 & 0.26  && 0.50  & 0.50 &   0.50 & 0.50 \\

Dynamic (1 violation)   & 0.46 & 0.48 & 0.48 & 0.48  && 0.50  & 0.50 &   0.50 & 0.50 \\

Dynamic (2 violations)  & 0.46 & 0.47 & 0.48 & 0.47  && 0.50  & 0.50 &   0.50 & 0.50 \\

\midrule
Total                   & 0.38 & 0.42 & 0.42 & 0.40  && 0.50  & 0.50 &   0.50 & 0.50 \\

\bottomrule
\end{tabular}
\end{table*}

\begin{table*}[h]
\centering
\caption{ Block O2 | Model: GAN (short-term prediction task) | Relative classification ($L_R$)}
\label{tab:modelRel}
\begin{tabular}{lccccccccc}
\toprule
& \multicolumn{3}{c}{Visible}& & &\multicolumn{3}{c}{Occluded}& \\ 

\cmidrule{2-4} \cmidrule{7-9}
Type of scene           &1 obj.&2 obj.&3 obj.& Total  &&1 obj. &2 obj.& 3 obj. & Total \\

\midrule
Static                  & 0.00 & 0.00 & 0.00 & 0.00  && 0.54  & 0.55 &   0.47 & 0.52 \\

Dynamic (1 violation)   & 0.44 & 0.38 & 0.47 & 0.43  && 0.56  & 0.52 &   0.54 & 0.54 \\

Dynamic (2 violations)  & 0.38 & 0.52 & 0.51 & 0.47  && 0.50  & 0.50 &   0.48 & 0.49 \\

\midrule
Total                   & 0.27 & 0.30 & 0.33 & 0.30  && 0.53  & 0.52 &   0.50 & 0.52 \\

\bottomrule
\end{tabular}
\end{table*}

\begin{table*}[h]
\centering
\caption{ Block O2 | Model: GAN (short-term prediction task) | Absolute classification ($L_A$)}
\label{tab:modelRel}
\begin{tabular}{lccccccccc}
\toprule
& \multicolumn{3}{c}{Visible}& & &\multicolumn{3}{c}{Occluded}& \\ 

\cmidrule{2-4} \cmidrule{7-9}
Type of scene           &1 obj.&2 obj.&3 obj.& Total  &&1 obj. &2 obj.& 3 obj. & Total \\

\midrule
Static                  & 0.29 & 0.30 & 0.32 & 0.30  && 0.50  & 0.50 &   0.50 & 0.50 \\

Dynamic (1 violation)   & 0.49 & 0.49 & 0.49 & 0.49  && 0.50  & 0.50 &   0.50 & 0.50 \\

Dynamic (2 violations)  & 0.48 & 0.49 & 0.50 & 0.49  && 0.50  & 0.50 &   0.50 & 0.50 \\

\midrule
Total                   & 0.42 & 0.43 & 0.44 & 0.43  && 0.50  & 0.50 &   0.50 & 0.50 \\

\bottomrule
\end{tabular}
\end{table*}

\begin{table*}[h]
\centering
\caption{ Block O2 | Model: CNN (long-term prediction task) | Relative classification ($L_R$)}
\label{tab:modelRel}
\begin{tabular}{lccccccccc}
\toprule
& \multicolumn{3}{c}{Visible}& & &\multicolumn{3}{c}{Occluded}& \\ 

\cmidrule{2-4} \cmidrule{7-9}
Type of scene           &1 obj.&2 obj.&3 obj.& Total  &&1 obj. &2 obj.& 3 obj. & Total \\

\midrule
Static                  & 0.00 & 0.00 & 0.01 & 0.00  && 0.50  & 0.47 &   0.52 & 0.50 \\

Dynamic (1 violation)   & 0.13 & 0.22 & 0.25 & 0.20  && 0.51  & 0.50 &   0.55 & 0.52 \\

Dynamic (2 violations)  & 0.11 & 0.10 & 0.17 & 0.13  && 0.56  & 0.49 &   0.53 & 0.53 \\

\midrule
Total                   & 0.08 & 0.11 & 0.14 & 0.11  && 0.52  & 0.49 &   0.54 & 0.52 \\

\bottomrule
\end{tabular}
\end{table*}

\begin{table*}[h]
\centering
\caption{ Block O2 | Model: CNN (long-term prediction task) | Absolute classification ($L_A$)}
\label{tab:modelRel}
\begin{tabular}{lccccccccc}
\toprule
& \multicolumn{3}{c}{Visible}& & &\multicolumn{3}{c}{Occluded}& \\ 

\cmidrule{2-4} \cmidrule{7-9}
Type of scene           &1 obj.&2 obj.&3 obj.& Total  &&1 obj. &2 obj.& 3 obj. & Total \\

\midrule
Static                  & 0.34 & 0.41 & 0.40 & 0.38  && 0.50  & 0.50 &   0.50 & 0.50 \\

Dynamic (1 violation)   & 0.43 & 0.45 & 0.46 & 0.45  && 0.50  & 0.50 &   0.50 & 0.50 \\

Dynamic (2 violations)  & 0.43 & 0.44 & 0.46 & 0.45  && 0.50  & 0.50 &   0.50 & 0.50 \\

\midrule
Total                   & 0.40 & 0.43 & 0.44 & 0.43  && 0.50  & 0.50 &   0.50 & 0.50 \\

\bottomrule
\end{tabular}
\end{table*}

\begin{table*}[h]
\centering
\caption{ Block O2 | Model: GAN (long-term prediction task) | Relative classification ($L_R$)}
\label{tab:modelRel}
\begin{tabular}{lccccccccc}
\toprule
& \multicolumn{3}{c}{Visible}& & &\multicolumn{3}{c}{Occluded}& \\ 

\cmidrule{2-4} \cmidrule{7-9}
Type of scene           &1 obj.&2 obj.&3 obj.& Total  &&1 obj. &2 obj.& 3 obj. & Total \\

\midrule
Static                  & 0.02 & 0.02 & 0.00 & 0.01  && 0.49  & 0.55 &   0.53 & 0.52 \\

Dynamic (1 violation)   & 0.35 & 0.42 & 0.54 & 0.44  && 0.50  & 0.40 &   0.45 & 0.45 \\

Dynamic (2 violations)  & 0.44 & 0.51 & 0.53 & 0.50  && 0.56  & 0.44 &   0.53 & 0.51 \\

\midrule
Total                   & 0.27 & 0.32 & 0.36 & 0.31  && 0.52  & 0.46 &   0.51 & 0.50 \\

\bottomrule
\end{tabular}
\end{table*}

\begin{table*}[h]
\centering
\caption{ Block O2 | Model: GAN (long-term prediction task) | Absolute classification ($L_A$)}
\label{tab:modelRel}
\begin{tabular}{lccccccccc}
\toprule
& \multicolumn{3}{c}{Visible}& & &\multicolumn{3}{c}{Occluded}& \\ 

\cmidrule{2-4} \cmidrule{7-9}
Type of scene           &1 obj.&2 obj.&3 obj.& Total  &&1 obj. &2 obj.& 3 obj. & Total \\

\midrule
Static                  & 0.40 & 0.39 & 0.38 & 0.39  && 0.50  & 0.50 &   0.50 & 0.50 \\

Dynamic (1 violation)   & 0.47 & 0.49 & 0.50 & 0.49  && 0.50  & 0.49 &   0.50 & 0.50 \\

Dynamic (2 violations)  & 0.47 & 0.50 & 0.50 & 0.49  && 0.50  & 0.50 &   0.50 & 0.50 \\

\midrule
Total                   & 0.45 & 0.46 & 0.46 & 0.46  && 0.50  & 0.50 &   0.50 & 0.50 \\

\bottomrule
\end{tabular}
\end{table*}

\begin{table*}[h]
\centering
\caption{ Block O2 | Model: CNN aggregated | Relative classification ($L_R$)}
\label{tab:modelRel}
\begin{tabular}{lccccccccc}
\toprule
& \multicolumn{3}{c}{Visible}& & &\multicolumn{3}{c}{Occluded}& \\ 

\cmidrule{2-4} \cmidrule{7-9}
Type of scene           &1 obj.&2 obj.&3 obj.& Total  &&1 obj. &2 obj.& 3 obj. & Total \\

\midrule
Static                  & 0.00 & 0.00 & 0.01 & 0.00  && 0.50  & 0.47 &   0.52 & 0.50 \\

Dynamic (1 violation)   & 0.13 & 0.22 & 0.25 & 0.20  && 0.51  & 0.50 &   0.55 & 0.52 \\

Dynamic (2 violations)  & 0.11 & 0.10 & 0.17 & 0.13  && 0.56  & 0.49 &   0.53 & 0.53 \\

\midrule
Total                   & 0.08 & 0.11 & 0.14 & 0.11  && 0.52  & 0.49 &   0.54 & 0.52 \\

\bottomrule
\end{tabular}
\end{table*}

\begin{table*}[h]
\centering
\caption{ Block O2 | Model: CNN aggregated | Absolute classification ($L_A$)}
\label{tab:modelRel}
\begin{tabular}{lccccccccc}
\toprule
& \multicolumn{3}{c}{Visible}& & &\multicolumn{3}{c}{Occluded}& \\ 

\cmidrule{2-4} \cmidrule{7-9}
Type of scene           &1 obj.&2 obj.&3 obj.& Total  &&1 obj. &2 obj.& 3 obj. & Total \\

\midrule
Static                  & 0.34 & 0.41 & 0.40 & 0.38  && 0.50  & 0.50 &   0.50 & 0.50 \\

Dynamic (1 violation)   & 0.43 & 0.45 & 0.46 & 0.45  && 0.50  & 0.50 &   0.50 & 0.50 \\

Dynamic (2 violations)  & 0.43 & 0.44 & 0.46 & 0.45  && 0.50  & 0.50 &   0.50 & 0.50 \\

\midrule
Total                   & 0.40 & 0.43 & 0.44 & 0.43  && 0.50  & 0.50 &   0.50 & 0.50 \\

\bottomrule
\end{tabular}
\end{table*}

\begin{table*}[h]
\centering
\caption{ Block O3 | Model: CNN (short-term prediction task) | Relative classification ($L_R$)}
\label{tab:modelRel}
\begin{tabular}{lccccccccc}
\toprule
& \multicolumn{3}{c}{Visible}& & &\multicolumn{3}{c}{Occluded}& \\ 

\cmidrule{2-4} \cmidrule{7-9}
Type of scene           &1 obj.&2 obj.&3 obj.& Total  &&1 obj. &2 obj.& 3 obj. & Total \\

\midrule
Static                  & 0.00 & 0.00 & 0.00 & 0.00  && 0.48  & 0.43 &   0.46 & 0.46 \\

Dynamic (1 violation)   & 0.34 & 0.38 & 0.46 & 0.39  && 0.47  & 0.48 &   0.50 & 0.49 \\

Dynamic (2 violations)  & 0.47 & 0.45 & 0.45 & 0.45  && 0.52  & 0.51 &   0.53 & 0.52 \\

\midrule
Total                   & 0.27 & 0.27 & 0.30 & 0.28  && 0.49  & 0.47 &   0.49 & 0.49 \\

\bottomrule
\end{tabular}
\end{table*}

\begin{table*}[h]
\centering
\caption{ Block O3 | Model: CNN (short-term prediction task) | Absolute classification ($L_A$)}
\label{tab:modelRel}
\begin{tabular}{lccccccccc}
\toprule
& \multicolumn{3}{c}{Visible}& & &\multicolumn{3}{c}{Occluded}& \\ 

\cmidrule{2-4} \cmidrule{7-9}
Type of scene           &1 obj.&2 obj.&3 obj.& Total  &&1 obj. &2 obj.& 3 obj. & Total \\

\midrule
Static                  & 0.22 & 0.21 & 0.23 & 0.22  && 0.50  & 0.49 &   0.50 & 0.50\\

Dynamic (1 violation)   & 0.49 & 0.49 & 0.49 & 0.49  && 0.50  & 0.50 &   0.50 & 0.50 \\

Dynamic (2 violations)  & 0.49 & 0.49 & 0.50 & 0.49  && 0.50  & 0.50 &   0.50 & 0.50 \\

\midrule
Total                   & 0.40 & 0.40 & 0.41 & 0.40  && 0.50  & 0.50 &   0.50 & 0.50 \\

\bottomrule
\end{tabular}
\end{table*}

\begin{table*}[h]
\centering
\caption{ Block O3 | Model: GAN (short-term prediction task) | Relative classification ($L_R$)}
\label{tab:modelRel}
\begin{tabular}{lccccccccc}
\toprule
& \multicolumn{3}{c}{Visible}& & &\multicolumn{3}{c}{Occluded}& \\ 

\cmidrule{2-4} \cmidrule{7-9}
Type of scene           &1 obj.&2 obj.&3 obj.& Total  &&1 obj. &2 obj.& 3 obj. & Total \\

\midrule
Static                  & 0.00 & 0.00 & 0.00 & 0.00  && 0.47  & 0.43 &   0.37 & 0.43\\

Dynamic (1 violation)   & 0.31 & 0.45 & 0.43 & 0.40  && 0.50  & 0.47 &   0.54 & 0.50 \\

Dynamic (2 violations)  & 0.34 & 0.42 & 0.43 & 0.40  && 0.48  & 0.52 &   0.54 & 0.51 \\

\midrule
Total                   & 0.22 & 0.29 & 0.29 & 0.26  && 0.48  & 0.48 &   0.48 & 0.48 \\

\bottomrule
\end{tabular}
\end{table*}

\begin{table*}[h]
\centering
\caption{ Block O3 | Model: GAN (short-term prediction task) | Absolute classification ($L_A$)}
\label{tab:modelRel}
\begin{tabular}{lccccccccc}
\toprule
& \multicolumn{3}{c}{Visible}& & &\multicolumn{3}{c}{Occluded}& \\ 

\cmidrule{2-4} \cmidrule{7-9}
Type of scene           &1 obj.&2 obj.&3 obj.& Total  &&1 obj. &2 obj.& 3 obj. & Total \\

\midrule
Static                  & 0.29 & 0.33 & 0.30 & 0.31  && 0.50  & 0.49 &   0.50 & 0.50\\

Dynamic (1 violation)   & 0.46 & 0.49 & 0.49 & 0.48  && 0.50  & 0.50 &   0.51 & 0.50 \\

Dynamic (2 violations)  & 0.44 & 0.47 & 0.47 & 0.46  && 0.50  & 0.50 &   0.50 & 0.50 \\

\midrule
Total                   & 0.40 & 0.43 & 0.42 & 0.42  && 0.50  & 0.50 &   0.50 & 0.50 \\

\bottomrule
\end{tabular}
\end{table*}

\begin{table*}[h]
\centering
\caption{ Block O3 | Model: CNN (long-term prediction task) | Relative classification ($L_R$)}
\label{tab:modelRel}
\begin{tabular}{lccccccccc}
\toprule
& \multicolumn{3}{c}{Visible}& & &\multicolumn{3}{c}{Occluded}& \\ 

\cmidrule{2-4} \cmidrule{7-9}
Type of scene           &1 obj.&2 obj.&3 obj.& Total  &&1 obj. &2 obj.& 3 obj. & Total \\

\midrule
Static                  & 0.02 & 0.00 & 0.00 & 0.01  && 0.48  & 0.49 &   0.47 & 0.48\\

Dynamic (1 violation)   & 0.45 & 0.52 & 0.43 & 0.47  && 0.54  & 0.48 &   0.53 & 0.52 \\

Dynamic (2 violations)  & 0.56 & 0.45 & 0.44 & 0.48  && 0.51  & 0.59 &   0.52 & 0.54 \\

\midrule
Total                   & 0.35 & 0.32 & 0.29 & 0.32  && 0.51  & 0.52 &   0.51 & 0.51 \\

\bottomrule
\end{tabular}
\end{table*}

\begin{table*}[h]
\centering
\caption{ Block O3 | Model: CNN (long-term prediction task) | Absolute classification ($L_A$)}
\label{tab:modelRel}
\begin{tabular}{lccccccccc}
\toprule
& \multicolumn{3}{c}{Visible}& & &\multicolumn{3}{c}{Occluded}& \\ 

\cmidrule{2-4} \cmidrule{7-9}
Type of scene           &1 obj.&2 obj.&3 obj.& Total  &&1 obj. &2 obj.& 3 obj. & Total \\

\midrule
Static                  & 0.35 & 0.36 & 0.40 & 0.37  && 0.50  & 0.50 &   0.50 & 0.50\\

Dynamic (1 violation)   & 0.50 & 0.50 & 0.50 & 0.50  && 0.50  & 0.50 &   0.50 & 0.50 \\

Dynamic (2 violations)  & 0.51 & 0.50 & 0.50 & 0.50  && 0.50  & 0.50 &   0.50 & 0.50 \\

\midrule
Total                   & 0.45 & 0.45 & 0.47 & 0.46  && 0.50  & 0.50 &   0.50 & 0.50 \\

\bottomrule
\end{tabular}
\end{table*}

\begin{table*}[h]
\centering
\caption{ Block O3 | Model: GAN (long-term prediction task) | Relative classification ($L_R$)}
\label{tab:modelRel}
\begin{tabular}{lccccccccc}
\toprule
& \multicolumn{3}{c}{Visible}& & &\multicolumn{3}{c}{Occluded}& \\ 

\cmidrule{2-4} \cmidrule{7-9}
Type of scene           &1 obj.&2 obj.&3 obj.& Total  &&1 obj. &2 obj.& 3 obj. & Total \\

\midrule
Static                  & 0.01 & 0.00 & 0.00 & 0.00  && 0.53  & 0.53 &   0.59 & 0.55\\

Dynamic (1 violation)   & 0.53 & 0.50 & 0.60 & 0.54  && 0.55  & 0.55 &   0.48 & 0.53 \\

Dynamic (2 violations)  & 0.42 & 0.51 & 0.54 & 0.49  && 0.43  & 0.52 &   0.52 & 0.49 \\

\midrule
Total                   & 0.32 & 0.34 & 0.38 & 0.34  && 0.50  & 0.53 &   0.53 & 0.52 \\

\bottomrule
\end{tabular}
\end{table*}

\begin{table*}[h]
\centering
\caption{ Block O3 | Model: GAN (long-term prediction task) | Absolute classification ($L_A$)}
\label{tab:modelRel}
\begin{tabular}{lccccccccc}
\toprule
& \multicolumn{3}{c}{Visible}& & &\multicolumn{3}{c}{Occluded}& \\ 

\cmidrule{2-4} \cmidrule{7-9}
Type of scene           &1 obj.&2 obj.&3 obj.& Total  &&1 obj. &2 obj.& 3 obj. & Total \\

\midrule
Static                  & 0.29 & 0.33 & 0.35 & 0.32  && 0.50  & 0.50 &   0.51 & 0.50\\

Dynamic (1 violation)   & 0.50 & 0.49 & 0.52 & 0.50  && 0.50  & 0.51 &   0.50 & 0.50 \\

Dynamic (2 violations)  & 0.50 & 0.49 & 0.50 & 0.50  && 0.50  & 0.50 &   0.50 & 0.50 \\

\midrule
Total                   & 0.43 & 0.44 & 0.46 & 0.44  && 0.50  & 0.50 &   0.50 & 0.50 \\

\bottomrule
\end{tabular}
\end{table*}

\begin{table*}[h]
\centering
\caption{ Block O3 | Model: CNN aggregated | Relative classification ($L_R$)}
\label{tab:modelRel}
\begin{tabular}{lccccccccc}
\toprule
& \multicolumn{3}{c}{Visible}& & &\multicolumn{3}{c}{Occluded}& \\ 

\cmidrule{2-4} \cmidrule{7-9}
Type of scene           &1 obj.&2 obj.&3 obj.& Total  &&1 obj. &2 obj.& 3 obj. & Total \\

\midrule
Static                  & 0.02 & 0.00 & 0.00 & 0.01  && 0.48  & 0.49 &   0.47 & 0.48\\

Dynamic (1 violation)   & 0.45 & 0.52 & 0.43 & 0.47  && 0.54  & 0.48 &   0.53 & 0.52 \\

Dynamic (2 violations)  & 0.56 & 0.45 & 0.44 & 0.48  && 0.51  & 0.59 &   0.52 & 0.54 \\

\midrule
Total                   & 0.35 & 0.32 & 0.29 & 0.32  && 0.51  & 0.52 &   0.51 & 0.51 \\

\bottomrule
\end{tabular}
\end{table*}

\begin{table*}[h]
\centering
\caption{ Block O3 | Model: CNN aggregated | Absolute classification ($L_A$)}
\label{tab:modelRel}
\begin{tabular}{lccccccccc}
\toprule
& \multicolumn{3}{c}{Visible}& & &\multicolumn{3}{c}{Occluded}& \\ 

\cmidrule{2-4} \cmidrule{7-9}
Type of scene           &1 obj.&2 obj.&3 obj.& Total  &&1 obj. &2 obj.& 3 obj. & Total \\

\midrule
Static                  & 0.35 & 0.36 & 0.40 & 0.37  && 0.50  & 0.50 &   0.50 & 0.50\\

Dynamic (1 violation)   & 0.50 & 0.50 & 0.50 & 0.50  && 0.50  & 0.50 &   0.50 & 0.50 \\

Dynamic (2 violations)  & 0.51 & 0.50 & 0.50 & 0.50  && 0.50  & 0.50 &   0.50 & 0.50 \\

\midrule
Total                   & 0.45 & 0.45 & 0.47 & 0.46  && 0.50  & 0.50 &   0.50 & 0.50 \\

\bottomrule
\end{tabular}
\end{table*}

\end{small}

\clearpage

\section{Human results (detailed)}

\begin{small}
\begin{table*}[h]
\centering
\caption{ Block O1 | Human evaluation |  Relative classification ($L_R$)}
\label{tab:modelRel}
\begin{tabular}{lccccccccc}
\toprule
& \multicolumn{3}{c}{Visible}& & &\multicolumn{3}{c}{Occluded}& \\

\cmidrule{2-4} \cmidrule{7-9}
Type of scene           &1 obj.&2 obj.&3 obj.& Total  &&1 obj. &2 obj.& 3 obj. & Total  \\

\midrule
Static                  & 0.01 & 0.06 & 0.00 & 0.02  && 0.12  & 0.22 &   0.20 & 0.18 \\

Dynamic (1 violation)   & 0.04 & 0.19 & 0.18 & 0.14  && 0.06  & 0.12 &   0.17 & 0.12  \\

Dynamic (2 violations)  & 0.04 & 0.25 & 0.09 & 0.13  && 0.26  & 0.10 &   0.13 & 0.16  \\

\midrule
Total                   & 0.03 & 0.17 & 0.09 & 0.10  && 0.15  & 0.15 &   0.17 & 0.15  \\

\bottomrule
\end{tabular}
\end{table*}

\begin{table*}[h]
\centering
\caption{ Block O2 | Human evaluation |  Relative classification ($L_R$)}
\label{tab:modelRel}
\begin{tabular}{lccccccccc}
\toprule
& \multicolumn{3}{c}{Visible}& & &\multicolumn{3}{c}{Occluded}& \\

\cmidrule{2-4} \cmidrule{7-9}
Type of scene           &1 obj.&2 obj.&3 obj.& Total  &&1 obj. &2 obj.& 3 obj. & Total  \\

\midrule
Static                  & 0.00 & 0.03 & 0.02 & 0.02  && 0.14  & 0.18 &   0.17 & 0.16 \\

Dynamic (1 violation)   & 0.16 & 0.04 & 0.22 & 0.14  && 0.12  & 0.23 &   0.09 & 0.15   \\

Dynamic (2 violations)  & 0.17 & 0.25 & 0.33 & 0.25  && 0.20  & 0.23 &   0.18 & 0.20  \\

\midrule
Total                   & 0.11 & 0.11 & 0.19 & 0.14  && 0.15  & 0.21 &   0.15 & 0.17  \\

\bottomrule
\end{tabular}
\end{table*}

\begin{table*}[h]
\centering
\caption{ Block O3 | Human evaluation |  Relative classification ($L_R$)}
\label{tab:modelRel}
\begin{tabular}{lccccccccc}
\toprule
& \multicolumn{3}{c}{Visible}& & &\multicolumn{3}{c}{Occluded}& \\

\cmidrule{2-4} \cmidrule{7-9}
Type of scene           &1 obj.&2 obj.&3 obj.& Total  &&1 obj. &2 obj.& 3 obj. & Total  \\

\midrule
Static                  & 0.23 & 0.10 & 0.24 & 0.19  && 0.32  & 0.17 &   0.40 & 0.30 \\

Dynamic (1 violation)   & 0.24 & 0.29 & 0.32 & 0.28  && 0.44  & 0.60 &   0.50 & 0.51   \\

Dynamic (2 violations)  & 0.06 & 0.21 & 0.20 & 0.16  && 0.38  & 0.57 &   0.44 & 0.46 \\

\midrule
Total                   & 0.18 & 0.20 & 0.25 & 0.21  && 0.38  & 0.45 &   0.45 & 0.42 \\

\bottomrule
\end{tabular}
\end{table*}

\clearpage
\section*{Detailed mask predictor}

\begin{table}[h]
\caption{Mask predictor (9747011 parameters). BN stands for batch-normalization.} \label{tab:maskPredictor}
\begin{center}
	\begin{tabular}{|c|}
	\hline
	Input frame \\
	3 x 64 x 64\\
	\hline
	7 first layers of resnet-18 (pretrained, frozen weights)\\
	\hline
    Reshape 1 x 8192\\
	\hline
	FC 8192 $\rightarrow$ 128\\
	\hline
	FC 128 $\rightarrow$ 8192\\
	\hline
	Reshape 128 x 8 x 8\\
	\hline
	UpSamplingNearest(2), 3 x 3 Conv. 128 - 1 str., BN, ReLU\\
	\hline
	UpSamplingNearest(2), 3 x 3 Conv. 64 - 1 str., BN, ReLU\\
	\hline
	UpSamplingNearest(2), 3 x 3 Conv. 3 - 1 str., BN, ReLU\\
	\hline
	3 sigmoid\\
	Target mask\\
	\hline
	\end{tabular}
\end{center}
\end{table}

\begin{figure}[h]
\begin{center}
   \includegraphics[width=0.4\linewidth]{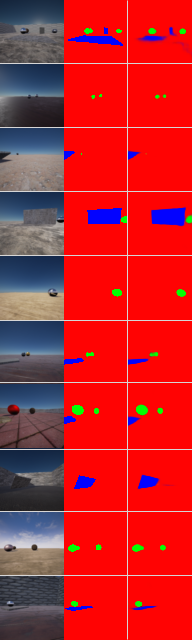}
\end{center}
   \caption{Output examples of our semantic mask predictor. From left to right: input image, ground truth semantic mask, predicted semantic mask.}
\label{fig:maskPred}
\end{figure}
\end{small}
\end{document}